\documentclass[nohyperref]{article}

\usepackage{microtype}
\usepackage{graphicx}
\graphicspath{ {./images/} }
\usepackage{booktabs} 
\usepackage{xcolor}
\usepackage{multirow}
\usepackage{caption}
\usepackage{subcaption}
\usepackage{algorithm}
\usepackage{hyperref}

\usepackage[accepted]{icml2022}

\usepackage{amsmath}
\usepackage{amssymb}
\usepackage{mathtools}
\usepackage{amsthm}
\usepackage{booktabs}
\usepackage{tabularx}
\usepackage{enumitem}
\usepackage{tabularx}
\usepackage{adjustbox}

\usepackage[capitalize,noabbrev]{cleveref}

\theoremstyle{plain}

\theoremstyle{definition}

\theoremstyle{remark}

\usepackage[textsize=tiny]{todonotes}

\begin{document}

\twocolumn[
\icmltitle{Causal Understanding For Video Question Answering}



\icmlsetsymbol{equal}{*}

\begin{icmlauthorlist}
\icmlauthor{Bhanu Prakash Reddy Guda}{equal,CMU}
\icmlauthor{Tanmay Kulkarni}{equal,CMU}
\icmlauthor{Adithya Sampath}{equal,CMU}
\icmlauthor{Swarnashree Mysore Sathyendra}{equal,CMU}
\icmlcorrespondingauthor{Tanmay Kulkarni}{tgkulkar@andrew.cmu.edu}
\end{icmlauthorlist}

\icmlaffiliation{CMU}{Carnegie Mellon University, Pittsburgh, USA}


\icmlkeywords{Machine Learning, ICML}

\vskip 0.3in
]


\printAffiliationsAndNotice{\icmlEqualContribution} 
\begin{abstract}
Video Question Answering is a challenging task, which requires the model to reason over multiple frames and understand the interaction between different objects to answer questions based on the context provided within the video, especially in datasets like NExT-QA \cite{nextQA} which emphasize on causal and temporal questions. Previous approaches leverage either sub-sampled information or causal intervention techniques along with complete video features to tackle the NExT-QA task. 
In this work we elicit the limitations of these approaches and propose solutions along four novel directions of improvements on the NExT-QA dataset. Our approaches attempts to compensate for the shortcomings in the previous works by systematically attacking each of these problems by smartly sampling frames, explicitly encoding actions and creating interventions that challenge the understanding of the model. Overall, for both single-frame (+6.3\%) and complete-video (+1.1\%) based approaches, we obtain the state-of-the-art results on NExT-QA dataset. 
   
\end{abstract}

\section{Introduction}
\label{sec:introduction}
Understanding video data brings us a step closer towards real world embodied agents. Previous challenges such as Visual Question Answering attempted to answer questions based on the information present in a single image or a scene. Videos add additional challenges with respect to understanding interactions across multiple frames, so as to reason about the situation being described. Not only do the agents need to understand events that have happened in previous frames, but they also need to reason about \textit{how} the events in the previous frames \textit{affect} the events happening in the current and/or future frames.

Current state of the art models such as VGT \cite{xiao2022video} attempt to connect relationships through associations between the video and text features through convoluted modules, which are prone to learning spurious correlations, and are often hard to debug. Another line of work \cite{Li_IGV, HGA+EIGV} attempts to take advantage of the invariance of the non-causal frames and equivariance of the causal frames as additional sources of supervision for enhancing the predictions on the data. However, both of these approaches seem to be lacking in generalization power, as they are only able to get test performance of $\approx50\%$ while human performance is around 90\%~\cite{nextQA}. 

On the other hand, approaches like ClipBERT~\cite{lei2021less} promote methods which rely on a subsampled information from the entire video to achieve compute tractability for VideoQA task. To this end, recently~\cite{buch2022revisiting} proposed the Atemporal Probe (ATP) approach for sampling the single most relevant frame for answering video-question pairs of VideoQA datasets MSR-VTT-MC~\cite{xu2016msr}, VALUE-How2QA~\cite{li2020hero,li2021value}. However, to achieve state-of-the-art performance in datasets like NExT~\cite{nextQA}, they concluded that entire video is required as the dataset predominantly consists of causal and temporal questions which require answering \textit{why \dots happened} or \textit{when \dots happened} in a given video.

In this work, we broadly categorize the prior works into \textit{single-frame} based (ClipBERT, ATP) and \textit{complete-video} based (IGV, EIGV, VGT) methods, and attempt to solve some of the severe limitations present in both categories of models. Motivated by the extensive analysis performed in our previous mid-term report, we propose to attempt the NExT-QA video question answering task, containing \textit{causal}, \textit{temporal} and \textit{descriptive} video question answer pairs, from four novel research directions. Extending prior art, we finally formulate the following four research questions (ideas), one for each direction, and aim to answer these through our proposed approaches.

\textbf{RQ1}: \textit{Instead of using either of the extremes, can we leverage a smart aggregation of the sub-sampled information to improve low compute approaches like ATP?}\\
\textbf{RQ2}: \textit{Can we improve the robustness of causal component based answering models like EIGV through hard-negative mining approaches, which could lead to performance boost?}\\
\textbf{RQ3}: \textit{Can we aim to improve grounding of video representations with the questions by extracting suitable information from videos like actions and descriptions?} \\
\textbf{RQ4}: \textit{Can we smartly identify the right frames/clips that need to be sampled in single-frame and alike low compute approaches for avoiding loss of information?}

In the quest of finding answers for these questions, we contribute the following to the research on VideoQA task. 
\begin{enumerate}[leftmargin=*,topsep=-3pt,itemsep=-3pt]
    \item Using only single-frame level compute, we propose \textbf{PCMA} model to bridge the gap between ATP and VGT baselines. In general, proposed PCMA is easy to transfer and replace multimodal fusion layers of any model.
    \item We propose to enhance video features by extracting and grounding with salient action and description information through \textbf{MAR} pipeline. These video features can be used with EIGV, ATP models with minimal changes.
    \item We highlight the drawbacks of using random clip based $do(.)$ for robustness and propose \textbf{MRI} pipeline with \textbf{MNSE} algorithm to generate hard-contrastive examples. The MRI also contributed to our final SOTA model.
\end{enumerate}
In addition to the obvious contributions, we also summarize the effectiveness of \textit{reinforcement learning} and \textit{teacher-student} paradigms for smartly sampling content without loss of information towards solving \textbf{\textit{RQ4}}. Overall we achieve a $\sim$6.3\% improvement in test accuracy for the single-frame models using a hybrid of the proposed PCMA and MAR techniques, and also achieve the state-of-the-art test performance (+1.1\%) on the NExT-QA dataset through the MAR and MRI techniques in the complete-video based approaches.  

\section{Related Work}
\label{related_work}

\textbf{Invariant Learning}
\label{invariant_learn_related_work}
\cite{NIPS1994_7fa732b5, sohn2012learning,  pmlr-v139-xiao21a, benton2020learning, Misra_2020_CVPR} show invariant learning(IL) as a paradigm to generalize better to data outside training distribution. 
For multi-modal tasks such as VideoQA, IL as means of visual grounding\cite{Li_IGV}, equivariant grounding when changes are made to casual/temporal scenes of videos\cite{li2022equivariant} and improving intrinsic interpreatibility have been explored on datasets like NExT-QA\cite{nextQA}.

\textbf{Contrastive Learning for better Interventions}
\label{mining_related_work}
We look at using constrastive loss for learning better intervention mechanisms. Current constrastive losses can be instance based - InfoNCE~\cite{oord2018representation}, UberNCE~\cite{han2020self}, MIL-NCE~\cite{miech2020end}; clustering based - CentreNCE~\cite{diba2021vi2clr}. ~\cite{dave2022tclr} propose a contrastive learning approach to preserve temporally varying video features using non-overlapping clips of the same video as negative samples. VQA tasks often rely on spurious relationships between language and images\cite{agrawal2016analyzing, zhang2016yin, johnson2017clevr}. This can be alleviated with contrastive loss approach proposed in \cite{li2022equivariant}, which we modify to provide a more robust approach for the VQA task.

\textbf{Grounded Video Representation and Aggregation}
\label{causal_learn_related_work}
Prior approaches~\cite{storks2019commonsense,wang2020causal,fang2020video2commonsense} attempt to extract causal knowledge using natural language. 
In  visual domain, causal reasoning has been used for downstream applications such as image classification~\cite{lopez2017discovering}, visual dialog~\cite{qi2020two}, image captioning~\cite{zhou2020more} and identifying hidden actions in videos~\cite{fire2017inferring}. 
Temporal reasoning has also been attempted within visual domain for understanding events in videos~\cite{zhou2018temporal,lin2019tsm,arnab2021vivit}. Multimodal approaches for this include Uniter~\cite{chen2020uniter}, Vilbert~\cite{lu2019vilbert}, Videobert~\cite{sun2019videobert} that use supervised/weakly-supervised methods of encoding visual-semantic knowledge into representations.

To the best of our knowledge, our proposed approaches of using cross modal grounded aggregation of frames from videos to minimize information loss in single-frame based approaches, enriching the intervention of videos by hard mining similar scenes, grounding videos by using salient action recognition and description modules, and smart-sampling of frames/clips from videos to maximize information gain are the first attempts towards enhancing VideoQA models through multiframe grounded approaches.


\section{Proposed Approach}
\label{proposed_approach}
In this section, we start by briefly describing the VideoQA task setup and then proceed to elaborate on various components of the proposed approaches, which we use towards answering our \textit{four research questions}. Given an input video $\mathcal{V}$, and a question $\mathcal{Q}$, and answer choices $\mathcal{C} = [\mathcal{C}_i]_{i=1}^{5}$, the objective is to minimize expected risk of 
using a multimodal framework $\mathcal{F}$ in predicting answer $\mathcal{A}\ \epsilon \ \mathcal{C}$ i.e., 
\begin{align}
min\ \mathcal{L}_{ERM}(\mathcal{F}(\mathcal{E_\text{v}(V); E_\text{t}(Q); E_\text{t}(C)}), \mathcal{A})
\label{eq:erm}
\end{align}
where $\mathcal{E_\text{v}}$, $\mathcal{E_\text{t}}$, $\mathcal{E_\text{t}}$ are the encoders that generate the representations for the $\mathcal{V}$, $\mathcal{Q}$, and $\mathcal{C}$ respectively. Often, we represent each of the choices individually i.e., $\mathcal{E_\text{t}(C) = [E_\text{t}(C_\text{i})]_{\text{i=1}}^{\text{5}}}$, and the encoders for the question and choices are same, as the modality of information for both of them is text (hence $\mathcal{E_\text{t}}$).

\subsection{Pairwise Cross Modal Aggregation (PCMA)}
\label{nc2}
\begin{figure*}[ht]
    \centering
    \includegraphics[width=0.7\textwidth]{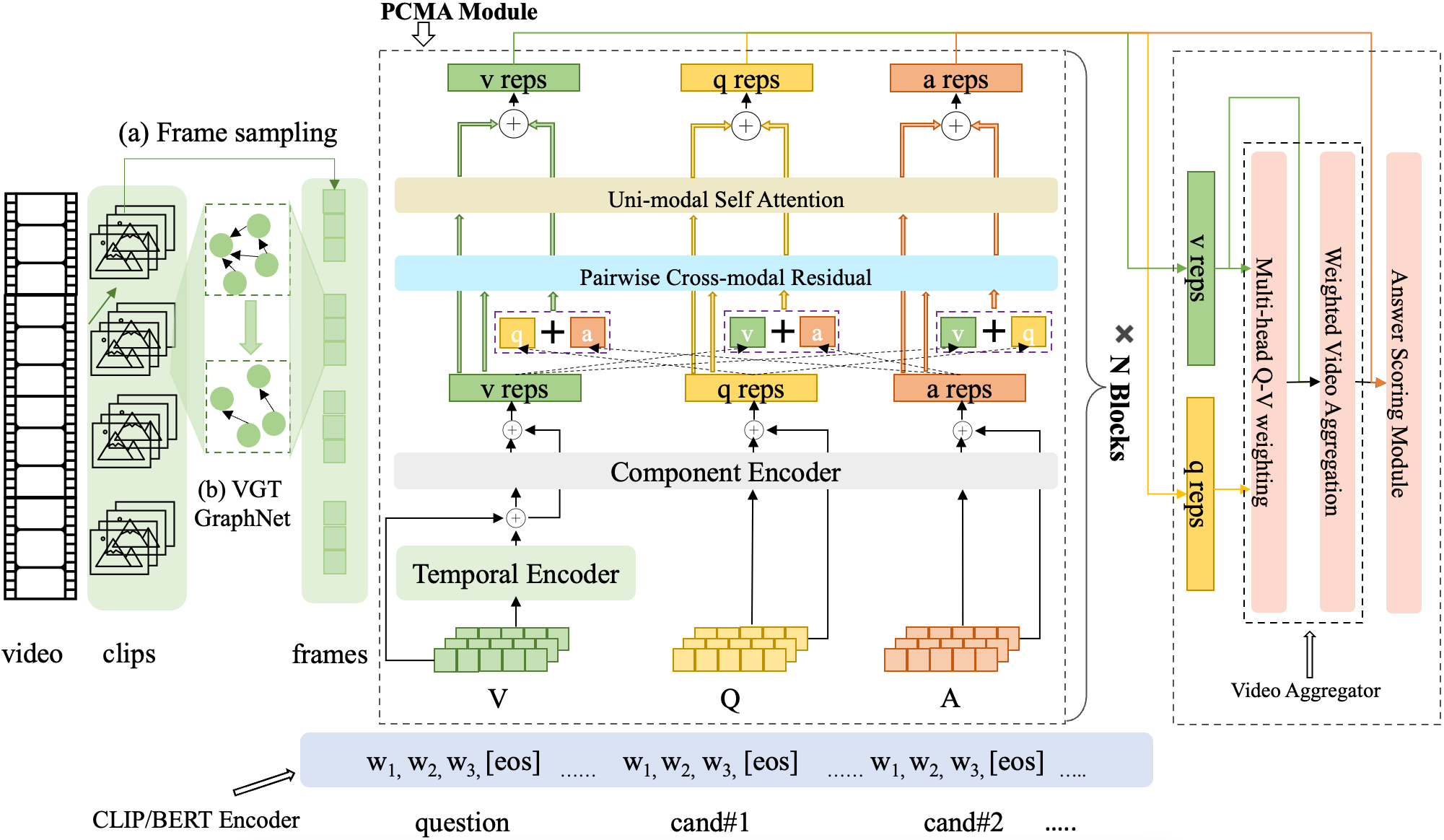}
    \caption{Model Architecture for the proposed PCMA model.}
    \label{fig:cma_block}
\end{figure*}
As highlighted in Section \ref{sec:introduction}, we can briefly categorize the VideoQA models into \textit{frame-based} and \textit{clip-based} methods. The components that we describe now are targeted at improving the \textit{less-compute} frame-based methods. Following~\cite{buch2022revisiting}, we use $\mathcal{N}$ frames, which are sampled uniformly or at random as inputs for VideoQA. While ATP uses all $\mathcal{N}$ frames as input, it makes use of a \textit{simple multimodal self-attention} based transformer to identify the single \textit{most relevant frame} $f$ for answering $\mathcal{Q}$ i.e., $ argmax_f P(f|\mathcal{Q}, \mathcal{V})$. We hypothesize that such extreme sampling could lead to severe loss of information, 
and to avoid this, we propose the PCMA model, which is presented in Figure~\ref{fig:cma_block}. The details of the novel components (middle section) of the architecture are as follows.\vspace{2mm}\\
\textbf{Temporal Encoder}:\quad Inspired by positional encoding used in the BERT model~\cite{devlin2018bert}, we leverage the timestamps of the video to encode the temporal information of the corresponding frame to better answer causal and temporal questions, which is absent in ATP and others. \\
\textbf{Component Encoder}:\quad Analogous to the segment encoders used in BERT and similar to transformer models, to distinguish between the different components, we propose to use a dense component encoder. \\
\textbf{Cross-modal Residual}:\quad Similar to the residual learning approach proposed in~\cite{residualmma}, we propose to use a pairwise cross modal attention mechanism, where each component acts as a query (Q) and uses rest of the components as the keys (K) and values (V) in cross attention module~\cite{vaswani2017attention}. The residual for a component is computed as 
\begin{align*}
    \small\text{MultiHead}\left[\text{softmax}\left(\frac{C1.[C2;C3]^\text{T}}{\sqrt{d_k}}\right)\left[C2;C3\right]\right]
\end{align*}
where $;$ is the concatenation operation and $C1, C2, C3$ are selected from $Combinations\{\mathcal{V}, \mathcal{Q}, \mathcal{A}\}$ components, and $d_k$ is the dimension of the component hidden representations. The learnt residual is then added to the self-attended component representations.\\
\textbf{Video Aggregator}($\mathbf{\mathcal{V}|\mathcal{Q}}$):\quad After $N$ PCMA blocks, we use the final $\mathcal{Q}$ representations and $\mathcal{V}$ representations to compute scores for the different parts of $\mathcal{V}$ (frames/clips) based on their relevance to $\mathcal{Q}$. The weighted aggregation for $\mathcal{V}$ is computed through a simple $\text{softmax}(\mathcal{Q}.\mathcal{V}^\text{T})$  operation. \\
\textbf{Answer Scorer}:\quad The final answer scoring module performs a cosine similarity operation followed by softmax layer using the final representations for $\mathcal{A}$ and the conditioned aggregated video representation $\mathbf{\mathcal{V}|\mathcal{Q}}$ to compute the logits for each if the choice in $\mathcal{C}$. 

We use the standard catergorical cross entropy as the $\mathcal{L}_{ERM}$ in Equation~\ref{eq:erm} using the above obtained logits. All the parameters of the PCMA modules are learnt end-to-end. Note that the parameters of the component specific encoders (e.g. CLIP/BERT/ResNet) are not trained as we focus on aligning the modality specific embeddings using multimodal fusion techniques. 

\subsection{Multimodal Action Grounding (MAG)}
\label{action_recognition}
\begin{figure*}[]
    \centering
    \includegraphics[width=1.0\textwidth,height=3.5cm]{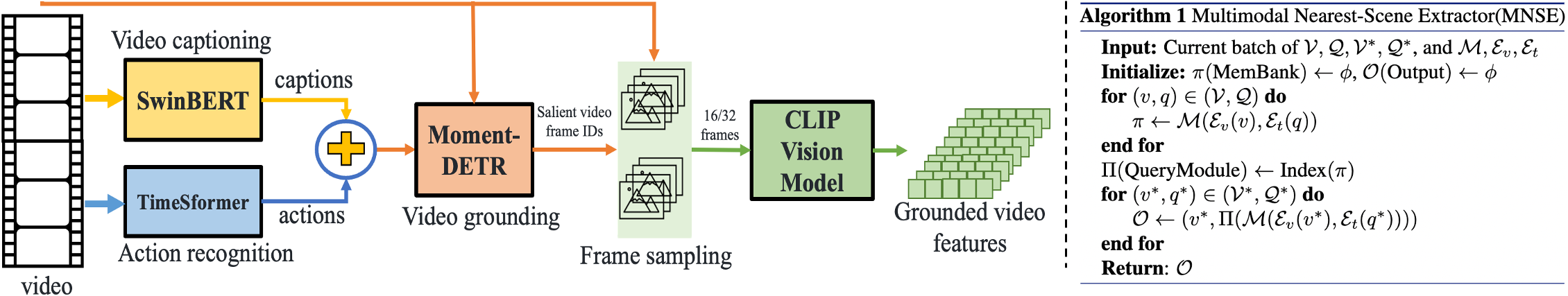}
    \caption{Left: Pipeline of the proposed MAG approach. Right: The MNSE algorithm used in the MRI approach}
    \label{fig:action_block}
\end{figure*}
In Section~\ref{related_work}, we summarized the prior art on video comprehension literature. Amongst these, action recognition in the videos is a non-trivial task which requires comprehending entire video information through a small number of abstract concepts. In this module, we emphasize particularly on action recognition and video description task as the NExT-QA dataset predominantly consists of descriptive, causal, and temporal questions.
The proposed Multimodal Action Grounding (MAG) module is also inspired by our human-in-the-box error analysis of causal and temporal questions, where we highlighted the importance of capturing actions taken by each entity in the video. The baseline models lack the strength for identifying \textit{when} and \textit{where} certain actions took place in the video, and we propose to provide this missing context. To achieve this goal, we propose an action grounding pipeline (reported in Figure~\ref{fig:action_block}-left), whose components are described here. 

\textbf{Action Recognition}:\quad The first stage of the pipeline is to detect the different actions that are performed by the entities in the video. Following \cite{gberta_2021_ICML}, we also model the action recognition as a high level video classification task. While the granularity of the predicted actions might be restricted at an high-level interaction as shown in examples like Figure~\ref{fig:action_qual_eg2}, this under-representation is solved in the next stage of description generation. \\
\textbf{Video Description}:\quad By leveraging pretrained object detection and interaction based description generation models \cite{lin2021end-to-end,2020mmaction2}, for each video in the NExT-QA dataset, we obtain the video descriptions and supplement with the action label for grounding the video in the language domain. \\
\textbf{Video Grounding}:\quad In the next stage of MAG, we concatenate the action labels with video descriptions to feed as language input for multimodal video grounding. In particular, we propose to leverage salient moment detector models~\cite{lei2021detecting,qvhighlights} to extract the \textit{keypoints} of the video by grounding the video using the actions and descriptions in the language domain to enhance the performance in question categories like \textit{when} and \textit{why}. The video along with the keypoint information are then streamed to the next stage for generating the final grounded representations. \\
\textbf{Frame Selection and Video Feature Extraction}: \quad In addition to predicting the key moments from the video, the saliency detector models also outputs the saliency scores for different frames of the video. As we also aim to reduce the compute cost and time in our research direction \textbf{RQ1}, we propose to directly sample the salient frames. With a pre-fixed sampling rate $\mathcal{N}$, we split the video into $\mathcal{N}$ clips at the top $\mathcal{N}$ salient points, and randomly sample one/many frames from the clips. Finally, we obtain the video representations by using any standard image encoding model like CLIP~\cite{radford2021learning}.

The output representations from the MAG model are then replaced with clip features of any previously proposed VideoQA backbone methods like EIGV~\cite{li2022equivariant} or HGA~\cite{HGA} models to predict the final answer using multimodal approaches. However, one should note that the final $\mathcal{N}$ representations are still generated using the entire video content at different stages of the MAG module and hence we classify this as \textit{full-video} based approach.

\subsection{Multimodal Robust Intervener (MRI)}
\label{mining}
\cite{li2022equivariant} proposed the Equivariant and Invariant grounding approach (EIGV) to enhance the VideoQA task performance by guiding the model to leverage only the causal moments of the video ($C$), where the causal component for a video $\mathcal{V}$ is defined by the corresponding question $\mathcal{Q}$. The core principle of this approach is that any \textit{intervention} to the non-causal parts($T$) of the video, shouldn't result in any change in the final predicted answer $\hat{\mathcal{A}}$, and any change in the original question ($\mathcal{Q}$) and/or changes to the causal components $C$ of $\mathcal{V}$ should warrant for a change. The former principle is the concept of \emph{Invariance} to non-causal frames, which is first proposed in~\cite{Li_IGV}, and later enhanced with the latter principle of \emph{Equivariance} to casual frames and questions in ~\cite{li2022equivariant}.  More concretely, this approach introduces an equivariant transformation $T_c$ to $C$ and $\mathcal{Q}$, and expects proportionate change in prediction $\hat{\mathcal{A}}$, and invariant transformation $T_i$ on $T$ shouldn't trigger any variation in $\hat{\mathcal{A}}$.
\begin{align*}
    T_t(\hat{A}) &= \mathcal{F}_{\hat{A}}(T_t({C}), T_t({\hat{\mathcal{Q}}})) \\
    \hat{A} &= \mathcal{F}_{\hat{A}}(T_i({T}), \mathcal{Q})
\end{align*}
\textbf{Intervention Pipeline}:\quad Similar to the interventions performed in~\cite{li2022equivariant}, we perform a linear mixup of two videos ($\mathcal{V}$ and $\mathcal{V}'$) generating a final video $\mathcal{V}^*$ and perform perturbations on the mixed video, question and answer datapoints.
To achieve the equivariance-invariance principle, first a \textit{Scene Intervener} creates the tuple $\langle \mathcal{Q}^*, \mathcal{A}^*, \mathcal{C}^*, \mathcal{T}^*\rangle$ with a similar linear mixup principle for questions, answers, causal, and non-causal components from the original videos $\mathcal{V}$ and $\mathcal{V}^{'}$, and their questions and answers. In the next step, a \textit{Causal Disruptor} generates additional $\mathcal{A}^+$ and $\mathcal{A}^-$ examples using the mixup data to use a \textit{Contrastive Learning} objective for pushing $\mathcal{A}^+$ toward $\mathcal{A}$ and pulling $\mathcal{A}^-$ away from $\mathcal{A}$. The positive and negative examples are generated by perturbing $T^*$ and $C^*$ components of $\mathcal{V}^*$ respectively using \textit{random clips} from other videos stored in a memory bank $\pi$. 
For detailed equations of \textit{Scene Intervener}, \textit{Causal Disruptor}, and \textit{Contrastive Learning} objective, please refer to Appendix~\ref{app:eigv-eq}. \\
\textbf{Multimodal Nearest-Scene Intervention}:\quad We modify the above pipeline by adding an intermediate module for finding nearest-scene using the algorithm shown in Algorithm \ref{fig:action_block}. At each iteration in the training phase, we feed the MNSE algorithm with the original $\mathcal{V}, \mathcal{Q}$, the mixup $\mathcal{V}^*, \mathcal{Q}^*$, the visual and text encoders ($\mathcal{E}_v,\mathcal{E}_t$), and a multimodal grounding module $\mathcal{M}$. The output $\mathcal{O}$ contains the \textit{nearest scene} $s$ for each video in $\mathcal{V}$. When $s$ is replaced with $C^*$ and $T^*$, we obtain the new robust $\mathcal{A}^-$, and $\mathcal{A}^+$ datapoints. The purpose of MNSE is to 
generate robust perturbations which restrict the model in finding spurious correlations/shortcut methods to answer the question $\mathcal{Q}$.

\subsection{Smart Sub-part Sampler (S3)}
\label{smart_sampling}
In addition to the smart frame sampling technique used in the MAG approach, we also propose two more ways to perform a smart sampling of the entire video. 

\textbf{Teacher-Student Sampler}:\quad Given a question $\mathcal{Q}$, we score (\textit{student model}) each of the frame from pool of input video frames and select only the top $\mathcal{N}$ frames to feed as input to the PCMA model (teacher model), and design the following custom loss for training the student model using teacher loss.
\begin{align*}
    \mathcal{L}_{t} &= \mathcal{L}_{ce} = \text{ce loss of the teacher model}\\
    \mathcal{L}_f &= \left\{
                              \begin{array}{@{}ll@{}}
                                P_s(f|\mathcal{V},\mathcal{Q})\times L_{s}, & \text{if}\ f\in \text{top-}\mathcal{N}\text{ frames}\\
                                P_s(f|\mathcal{V},\mathcal{Q})\times \lambda, & \text{otherwise}
                              \end{array}\right.\\
    \mathcal{L}_s &= \text{Batch Mean}(\frac{1}{|\mathcal{V}|}\times \sum_{f} L_f)
\end{align*}
where $\mathcal{L}_s=\mathcal{L}_{ce}$ is the loss obtained from the teacher PCMA on a single datapoint, $L_f$ is the loss for a single frame in the video $\mathcal{V}$ and $\mathcal{L}_s$ is the final aggregated loss of the student model. As evident from the loss function, this model is targeted at improving the PCMA component by specifying what frames to be chosen from the video as the input for the cross modal attention and aggregation procedure.

\textbf{RL for VideoQA}:\quad On the contrary, instead of using a multi-stage approach, we also explore the possibility of reformulating the VideoQA as an end-to-end reinforcement learning (RL) problem. In this setting, an agent is designed to select the most relevant frames from the video given the question. 
Thus, at every \textit{state}, the \textit{agent} picks an \textit{action} to select or not-select a frame. Based on the frames selected by the agent and the input question, a prediction backbone attempts to predict the correct answer. Hence, we design the reward for the RL agent as the negative loss obtained at the end of each episode. To control the size of the video being used at the end, we also penalize the agent actions proportional to the number of frames selected by the agent as relevant. An episode is defined as the agent being able to view all the frames in the video. Since the reward is very sparse, we also condition the agent to differentiate between next frame and a random frame as the intermediate dense reward.

\section{Experimental Setup}
\label{methodology}

In this section, we elaborate on the dataset, inputs, architectural, and hyperparameter choices for the proposed novel components \textbf{PCMA}, \textbf{MAG}, \textbf{MRI}, and \textbf{S3}. We also briefly describe our various experimental settings which we designed to the four research directions through the four questions (\textbf{\textit{RQs}}) mentioned in Section~\ref{sec:introduction}. More nuanced changes to the experiments and detailed analysis of the effectiveness of these components are presented in the Results section~\ref{results_and_discussion}.
\subsection{Dataset}
For this work, we run all the approaches on the NExT-QA \cite{nextQA} dataset. It consists of 5,440 videos with an average duration length of 44s. There are a total of 52,000 question answer pairs divided into causal, temporal and descriptive types respectively. This dataset is challenging due to the fact that the model needs to perform both causal and temporal reasoning on the video frames in order to answer the questions. The \textit{input modalities} for the task are the visual modality from the videos and text modality from questions and answer choices.

\subsection{Multimodal Baselines}
\label{sec:baselines}
Now, we briefly describe three baseline methods which are the most recent state-of-the-art approaches that attempted to solve NExT-QA task through different hypotheses.

\textbf{Atemporal Probe:} \cite{ATP} It investigates the need for multiple frames to solve a task. This model attempts to pick the best frame from the video $\mathcal{V}$ for solving the question given the information about the questions $\mathcal{Q}$ and answer choices $\mathcal{C}$.  

\textbf{EIGV:} \cite{HGA+EIGV} This model takes a causal approach towards solving the problem by predicting the relevant ($C$) and irrelevant parts ($T$) of the video $\mathcal{V}$ and replacing them with clips from other videos to make sure that the model is invariant to a change in the irrelevant part and equivariant to a change in the relevant part.

\textbf{VGT:} \cite{xiao2022video} The most recently proposed VGT model identifies the relationship between objects obtained with the help of object detection. They then model the interactions by constructing a graph of objects and the spatial relations between them. Finally, they perform late multimodal fusion on this aggregated representation with the information from question and the answer.

We compare our proposed modules, PCMA with the ATP method (\textit{single-frame}), and MAG, MRI components against EIGV and VGT models (\textit{complete-video}). 
\subsection{Proposed Approach}
\label{main_exp_method}
For train, test and validation, we use the same splits as those provided by the dataset. In addition, the dataset also contains pre-computed features corresponding to video, questions and answers. The video features are obtained by first dividing the video into 16 clips of equal length and then running through 3D ResNeXt-101 \cite{hara3dcnns} to obtain embeddings of dimension \texttt{16 x 4096}. For text modality, the dataset also provides BERT \cite{devlin2018bert} embeddings of Question-Answer pairs of the form \texttt{QUESTION [SEP] ANSWER} which ~\cite{nextQA} have fine-tuned on the NExT dataset to generate embeddings of dimension \texttt{5 x 37 x 768}. 
For all our experiments, we follow a similar process of splitting the video into equal length clips and select one frame for single-frame methods and aggregate all frames for complete-video based methods. Unless stated otherwise, the $\mathcal{E}_v, \mathcal{E}_t$ are CLIP for ATP and PCMA, and ResNeXt-101, BERT for EIGV based methods. The details on encoding clips for VGT model are presented in Appendix~\ref{vgt}. The specific architecture choices made for the PCMA, MAG, MRI and S3 modules are as follows.

\textbf{PCMA}:\quad We sample the first frame from each of the clips, which is treated as the representative frame of the entire clip~\cite{buch2022revisiting}. The \texttt{512} dimensional CLIP~\cite{clip} representations of the frames and the question, answer choices are used as the inputs in the architecture shown in Figure~\ref{fig:cma_block}. For the cross modal and self modal attention and aggregation mechanisms, we leverage the Multiheaded Transformer~\cite{vaswani2017attention} units. For final prediction, we simply align answer choice representation with the aggregated video representation through cosine similarity. 

\textbf{MAG}:\quad The input video is processed at 1FPS due to computation limitations and fed as inputs for action recognition and textual description models. In this work, we use the pretrained TimeSformer \cite{gberta_2021_ICML} model trained on the Kinetics-400 dataset \cite{kay2017kinetics} for action recognition, and SwinBERT \cite{lin2021end-to-end} model, pretrained on the VATEX dataset \cite{wang2019vatex} to get video descriptions. We use the pretrained models and implementations from MMAction2 \cite{2020mmaction2} for both the tasks. For further grounding the video we first concatenate the embeddings corresponding to \texttt{ACTION x [SEP] x DESCRIPTION x [SEP] x QUESTION x [SEP] x ANSWER} to obtain \texttt{768} dimensional BERT \cite{devlin2018bert} features. The generated textual representation along with the sampled video are fed to the Moment DETR \cite{lei2021detecting} model pretrained on the QVHighlights dataset~\cite{qvhighlights}, for detecting the salient moments in the video $\mathcal{V}$. For sampling $\mathcal{N}$ salient frames, we experiment with two levels of sampling (1) MAR-16 - we first randomly sample 8 frames in the duration described by the moment coordinates with the highest saliency scores. Next, we divide the video into 4 equal segments, and randomly sample 2 frames from each section. We perform this additional sampling since we observed in our human-in-the-box analysis that some temporal questions ask questions related to what happens in the start/middle/end of the video (2) For MAR-32, we follow similar approach of 16 frames but with twice the values. The selected salient frames are encoded using the CLIP model.

\textbf{MRI}:\quad In the MNSE algorithm~\ref{fig:action_block}, for this work, we use a nearest neighbour approach for indexing and querying similar scenes. To handle nearest neighbour querying at large scale on the fly (for each batch), we implement the FAISS method proposed in \cite{johnson2019billion} for the NExT-QA dataset using the Hugging Face framework\footnote{\url{https://huggingface.co/course/chapter5/6}}. We experiment with three different ways of incorporating the MNSE into EIGV model (1) \textit{Static FAISS} (F1) - Populate the memory bank $\pi$ with all the video data and query for nearest scenes ($s$) at the start of the training process and keep them constant (2) \textit{Dynamic FAISS} (F2) - Perform a call to the MNSE algorithm at each batch to dynamically vary the scenes as the training progresses (3) \textit{Dynamic FAISS + $\pi^{{\mathcal{V}}^*}_d$} (F3) - Previous two setups append to $\pi$ for only one of $\mathcal{V}, \mathcal{V}^{'}$ (refer to Section \ref{mining}). In this setup, we extend the MNSE to accommodate the linear mixup videos $\mathcal{V}^*$ by populating $\pi$ with the corresponding mixup representations. Since we experiment with MRI on top of EIGV, the video representation dimension in the MNSE is of size {\tt 4096}.
\begin{table*}[ht]
    \begin{minipage}{.47\linewidth}
      \centering
        \begin{tabular}{clcccc}
            \toprule
                & Model & Acc\_C & Acc\_T & Acc\_D & Acc\\
            \midrule
                \multirow{12}{*}{\rotatebox{90}{$\longleftarrow$ Compute (extent of video)}} & ATP & 39.32	& 44.23	& 45.17	& 41.81 \\
                & PCMA & 44.38 & 42.99 & 61.65 & 46.27 \\
                & PCMA-80 & 46.07	& 45.22	& 62.68	& 48.38 \\
            \cline{2-6}
                & \small{PCMA+MAR-16} & 44.5 & 45.1 & 64.22 & 47.76 \\
                & \small{PCMA+MAR-32} & 44.8 & 45.78 & 63.96 & 48.1 \\
            \cline{2-6}
                & EIGV & 51.29  & 53.11  & 62.78  & 53.74\\
                & \small{+ MAR-16} & 52.13 & 53.29 & 63.99 & 54.44\\
                & \small{+ MAR-32} & 52.64 & 52.58 & \textbf{64.63} & 54.59\\
                & \small{+ F} & 51.09 & 51.49 & 60.36 & 52.73\\
                & \small{+ F + MAR-16} & 52.73 & 53.52 & 62.56 & 54.49 \\
                & \small{+ F + MAR-32} & \textbf{53.09} & \textbf{53.78} & 62.56 & \textbf{54.86}\\
                & VGT & 50.13& 52.85& 62.93& 53.0\\
            \bottomrule
        \end{tabular}
        \caption{Performance of different models on the test split. All the reported values correspond to the \% accuracy metric.}
        \label{tab:overall_results}
    \end{minipage}%
    \qquad
    \begin{minipage}{.47\linewidth}
      \centering
        \scalebox{0.8}{\begin{tabular}{lcccc}
            \toprule
            Model & Acc\_C & Acc\_T & Acc\_D & Acc\\ 
            \midrule
            EIGV + F1 & 51.20 & 52.43  & 61.64  & {53.29} \\ 
            EIGV + F2 & 50.93 & 51.79  & 61.57  & 52.94 \\ 
            EIGV + F3 & 51.09 & 51.49 & 60.36 & 52.73 \\ 
            \bottomrule
        \end{tabular}}
        \caption{Test performance of types of MRI}
        \label{tab:faiss_types}
~\\
        \scalebox{0.8}{\begin{tabular}{lcccc}
            \toprule
            Model & Acc\_C & Acc\_T & Acc\_D & Acc\\ 
            \midrule
            EIGV &  52.01 &  52.23  &  64.61   & {54.04} \\
             + F2 & 50.86 & 52.61  & 62.29  & 53.20 \\
            \midrule
            EIGV + seen-$do(.)$ & 48.14  & 50.62  & 58.69  & {50.58} \\ 
             + F2 + seen-$do(.)$ & 46.49 & 46.71  & 56.11  & 48.06 \\ 
            \midrule
            EIGV + unseen-$do(.)$  & 46.57 & 46.84  & 53.67  & 47.76  \\ 
             + F2 + unseen-$do(.)$ & 47.49 & 49.19  & 58.82  & {49.80} \\ 
            \bottomrule
        \end{tabular}}
        \caption{Val performance of ablation of MRI type 2 with EIGV}
        \label{tab:eigv_expts}
    \end{minipage} 
\end{table*}

\textbf{S3}:\quad The teacher-student variant for the S3 component follows a similar cross modal attention transformer for sampling the frames. At each layer, we learn $\mathbf{v} + \text{cross-attention}(\mathbf{v}, \mathbf{q})$, where $\mathbf{v}, \mathbf{q}$ are the video and question representations. The final stage in the student model is a linear layer followed by softmax to predict the probabilities $P_s(f|\mathcal{V},\mathcal{Q})$ (Section~\ref{smart_sampling}) for each frame $f$. In the RL method, we initialize the buffer of the agent with text embedding of the question. At each step, the state representation is obtained by taking the last state of the state model $\mathcal{F}_{state}$, which takes in frames selected by the model $\mathcal{V}_{buf}$ as the input. The agent based on the state representation and its own policy $\mathcal{F}_{policy}$ decides to add $\mathcal{V}_{frame}$. At the end of each episode, the sparse reward is obtained from the loss of the prediction backbone $\mathcal{F}_{pred}$ and the ratio of the selected frames to the total number of frames. The state model is defined as a simple transformer encoder, the policy network is a simple two layer MLP with standard VideoQA backbone.
\section{Results and Discussion}
\label{results_and_discussion}
In this section, we report the effectiveness of the proposed approach through several quantitative, qualitative, and ablation studies. The comprehensive set of results are reported in Table~\ref{tab:overall_results}. Overall, the proposed approach of using PCMA in single-frame based models outperforms ATP by huge margin, and the MAR + MRI improve the performance of complete-video based baseline EIGV by 1.12\%. 
\subsection{Effectiveness of PCMA Aggregation}
In Table~\ref{tab:overall_results}, the first part showcases the effectiveness of using PCMA aggregation over ATP's single-frame hypothesis. To reiterate, the improvements of PCMA over ATP are two-fold (1) explicit cross modal attention based residual learning to enhance uni-modal information, and (2) question conditioned video information aggregation instead of single-frame sampling. Though the PCMA uses an aggregation of $\mathcal{N}=16$ frames and ATP selects one out of 16, the compute resources are almost similar for both approaches since ATP also uses 16 frames to arrive at single frame. The performance boost of $\sim$4.4\% over ATP advocates for the use of PCMA\footnote{For qualitative example please refer to Figure~\ref{fig:cma_qual_eg}}. 

\textbf{Mitigating Sampling Bias}\quad As mentioned in Section~\ref{methodology}, the ATP and vanilla PCMA uses the first frame of each clip to represent the entire clip. However, this approach leads to sampling bias and hence lower generalization to test environment. We mitigate this by splitting the video into a higher number $\mathcal{N}=80$ clips, randomly sample one frame for each clip, and at each training iteration, sub-sample $\mathcal{N}=16$ frames from these 80 frames. We can observe a $\sim$2\% improvement in the test accuracy i.e acts as a regularizer against overfitting on sampling bias. 

With this analysis, we answer our first research question \textbf{\textit{RQ1}}: \textbf{\textit{RA1}} \textit{While optimizing for compute resources through sub-sampling is important, we can mitigate the tradeoff in performance to some extent through enhanced multimodal grounding and information aggregation approaches.} 

\subsection{MAR Module Contribution}
As the MAR module simply generates $\mathcal{N}=16$ length representations from a video $\mathcal{V}$, we need a backbone model to perform the downstream we replace the CLIP representations used in the PCMA model and incorporate the MAR module features. As observed from middle two rows of Table~\ref{tab:overall_results}, PCMA+MAR consistently improves over vanilla PCMA (except when using \textit{pseudo} 80 frames). MAR-16 and 32 correspond to two sampling strategies for summarizing $\mathcal{V}$ through actions as described in Section~\ref{methodology}.
\subsection{Improvement in Robustness through MRI}
To reiterate, the key principle of EIGV method is to correctly identify the causal and non-causal/complement scenes through $do(.)$ calculus operations like intervention and thereby enhance the overall performance of VideoQA task. However, in the baseline EIGV method, the $do(.)$ operations are performed using random clips, causing inconsistencies in the videos. Table~\ref{tab:faiss_types} reports the results from replacing the random $do(.)$ with MNSE-$do(.)$. Here F1, F2, F3 corresponds to the three types of MNSE as described in Section~\ref{main_exp_method}. We observe a consistent drop in performance with the three types of MSNE-$do(.)$ interventions. 

To understand this effect, we further experimented with seen vs unseen intervention evaluation setups and reported the results in Table~\ref{tab:eigv_expts}. The base models (top rows) for comparison are EIGV and EIGV + F2, i.e., MNSE-$do(.)$ with type F2 intervention. We limit our experiments to one type of intervention due to space and compute restrictions. From here on wards, all the interventions of MNSE are by default type 2, unless specified. Middle rows report the result in case of \textit{seen interventions}, where the inputs are the positive samples $\mathcal{A}^{+}$ from \textit{Causal Disruptor}, and bottom rows corresponds to the results of \textit{unseen interventions}, where we use MNSE-$do(.)$ to intervene and predict the answer using vanilla EIGV model and vice-versa. Since we are changning only non-causal parts, in all the cases, the expected answer remains unchanged, and hence the results are directly comparable to the top rows. 

Through these experiments, we observe that the drop in performance for vanilla EIGV on intervened instances is $\sim$3.5\% in the seen case and $\sim$6.3\% for unseen case. However, for the EIGV + F2 model, in both cases, the drop is consistent, in fact better in case of unseen intervention, which is the simple random intervention. We hypothesize that the results on simple random intervention are typically better than MSNE due to the abrupt discontinuities from random clips. Based on these results, we conclude that \textbf{\textit{RA2}} \textit{proposed MSNE based $do(.)$ interventions lead to a better robustness of the EIGV model by precisely identifying causal components and to some extent mitigate the abrupt discontinuities in intervened videos}, as the answer to our \textbf{\textit{RQ2}}. We see that the MSNE based MRI helps boost generalization and reduce overfitting to a single type of intervention. Qualitative examples for MRI are presented in Appendix Figures~\ref{fig:faiss_qual_eg1}\&\ref{fig:faiss_qual_eg2}
\begin{figure*}[ht]
    \centering
    \includegraphics[scale=0.35]{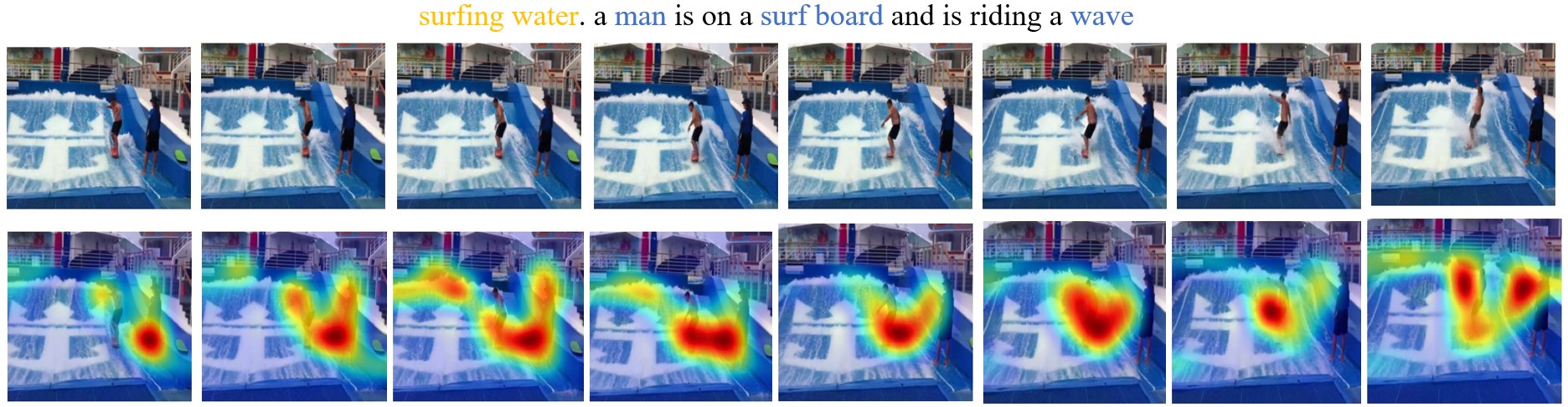}
    \caption{The model is able to match the most relevant frame with the relevant parts of the text}
    \label{fig:action_qual_eg1}
\end{figure*} 
\begin{figure}[]
    \centering
    \includegraphics[width=1.0\columnwidth]{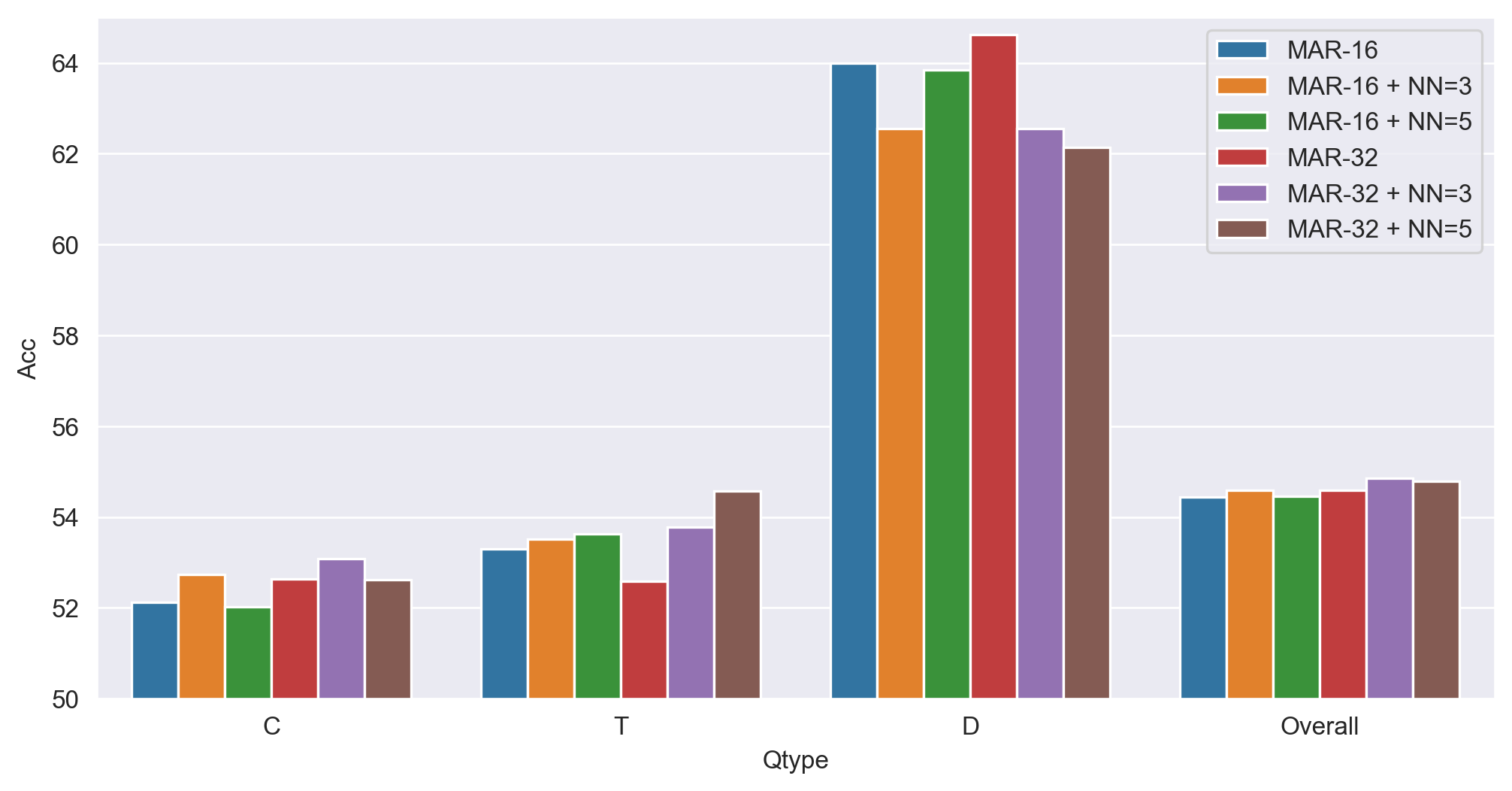}
    \caption{Test performance of the effect of the number of nearest neighbors selected usign the MNSE with MAR-16 and MAR-32 variants.}
    \label{fig:action_ground_acc}
\end{figure}
\subsection{Joining the components together}
Previously, in the MAR analysis part, we have discussed about the improvements contributed by the MAR representations over standard CLIP visual representations for the frames. Now we try to combine the MAR module with the vanilla and MNSE based MRI intervention methods. As observed from Table~\ref{tab:overall_results}, the best performing method is the MAR-32 module with EIGV baseline as as the VideoQA backbone, in the presence of robust interventions from the proposed MNSE algorithm. Note that we do not perform any joint modeling approaches between the PCMA and MRI because the intervention mechanism of the EIGV is not compatible with the cross modal and uni modal self attention interactions, and the changes that would make it possible are not trivial. However, since the MAG is just a novel method of generating action grounded video representations, it is compatible with both MRI and PCMA models. As mentioned several times earlier, the EIGV is a complete video based approach and ATP is a single-frame based approach. Following this notion, we have ordered the different models in Table~\ref{tab:overall_results} from top to down based on the model size and training time. 

As observed from Table~\ref{tab:overall_results}, there is a clear gap between the single-frame based and complete-video based methods. The proposed PCMA and MAR components helps to bridge this gap and the joint MAR +PCMA models lies in between the two extremes. The negative effect of adding MRI module to the vanilla EIGV is suppressed by the MAR module and hence led to the overall state-of-the-art result on the NExT-QA dataset. We hypothesize that identifying causal components just from the multimodal encoders $\mathcal{E}_v, \mathcal{E}_t$ is difficult compared to the MAR augmented representations where the actions are clearly encoded in the representations. A qualitative example depicting the salient regions of the images for generating the final CLIP representations is shown in Figure~\ref{fig:action_qual_eg1}.

Thus we answer our third question \textbf{\textit{RQ3}} as follows \textbf{\textit{RA3}} \textit{extracting information like actions, descriptions etc\dots, supplement the video representations with the help of grounding and hence improve the overall VideoQA task in both single-frame and complete-video based methods}. 

\textbf{Number of nearest neighbor scenes}\quad Instead of greedily choosing a single nearest neighbor scene for replacing causal ($C$) and non-causal ($T$) parts of $\mathcal{V}$, we also explored the effect of extracting a higher number of nearest neighbors and then randomly sampling one amongst them. To some extent the higher the number of nearest scenes, the harder is the intervention mechanism as randomness is added on top of nearest scene. The results are plotted in Figure~\ref{fig:action_ground_acc}. The key points to note here are: (1) Descriptive questions are more susceptible to the changes, and hence as the NN value increases the performance decreases in both MAR-16 and MAR-32 (2) This observation is exactly reverse for the temporal questions and (3) In causal question, the NN=3 works better than vanilla and NN=5 cases. This result showcases that it is not trivial to design a single best mechanism across all types of questions. 

\subsection{Issues with the VGT model}
Although VGT is the most recent best performing baseline proposed by \cite{xiao2022video}, we refrain from using it in our experiments and instead choose the next best baseline method EIGV~\cite{li2022equivariant}. Being easily transferable, we modified the multimodal fusion layers of VGT with those of the PCMA model, and to our surprise, we have observed a drastic drop in performance across all question types (by at least 6\%). Upon further experimentation, we found that removing the answer component from the PCMA's cross modal fusion led to a slight improvement of 0.22\% in the overall test accuracy (Figure~\ref{fig:cma_acc}). 

To understand this weird phenomenon, we have performed an extensive analysis of different components of the VGT model (refer to Appendix~\ref{vgt}). Unlike any other baseline or proposed model, only VGT tunes the 120M parameters of BERT. Even in the absence of question and any multimodal fusion layers, just by using cosine similarity between answer and video embeddings, we were able to achieve a similar accuracy of $\sim$53\% using VGT model. Hence we performed a qualitative analysis by rephrasing / rewording some of the answer correct answer choices and tried to retrieve the most similar answer choices from the entire set of answer choices. When rephrased the answers from {\tt look to her left} to {\tt turn to her left}, the model scored a higher similarity with the choice {\tt happy}, and similarly the choice {\tt resting} when phrased as {\tt taking rest}, the model's next highest similarity with the video was with the answer choice {\tt tigers}. Our hypothesis for this phenomenon is that by training the text embeddings along with the convoluted graph transformers, the VGT was capable enough to find shortcuts from the answer choices to video representations. However, this issue is not observed in any other model because we freeze the text and video representations and learn only the multimodal fusion layers/intervention layers. 

Finally, our experiments involving the final \textbf{\textit{RQ4}} didn't lead to any successful outcome. For the RL based VideoQA, we found that the reward signal was too weak even with the proxy dense rewards of predicting the next frame in the video sequence. Moreover the the underlying distribution for the prediction and state models are not stationary like any typical RL setup. Hence we observed a very noise learning curves in this setup. The teacher student model also suffered from noisy curves. On the other hand, we have observed a minor improvement of 0.6\% over the vanilla PCMA model. However, this model also quickly converges to a higher loss value even after tuning the hyperparameter $\lambda$ of the student loss function (Section~\ref{methodology}). Despite, the sampling technique used in the MAR module advocates for the improved performance through \textit{appropriate subsampling methods} (saliency based for PCMA+MAR) instead of random frames in the vanialla PCMA. Hence we conclude our analysis by providing a partly explored {\tt rephrased answer} for our \textbf{\textit{RQ4}}: \textbf{\textit{RA4}} \textit{smart subsampling techniques might be required to avoid loss of information, however, the subsampling models and loss functions themselves should be carefully tuned and therefore could be heavy in terms of runtime and compute}. 

\section{Conclusion and Future Work}
In this work, we established the need for solving the proposed four novel research questions and through extensive experiments, we have arrived at the answers for each of them. We showcased the fusion power of the PCMA modules for the NExT dataset and leveraged the action grounded video representations to enhance both the PCMA and EIGV+MRI models in single-frame and complete-video based methods. In addition to improved robustness, we have also observed that the MRI approach helps in achieving state-of-the-art performance on the NExT dataset. 

In future we intend to extend these components to the other VideoQA datasets. Our ablation study on the VGT baseline method advocates for the need of such extensive analyses in future to eliminate any spurious correlations. Our teacher-student model also acts as the proof-of-concept for the future research threads on developing smart sampling techniques.

\vspace{10pt}
\textbf{Github Repository}

\href{https://github.com/11777-MMML/11777-videoQA}{https://github.com/11777-MMML/11777-videoQA}

\bibliography{main}

\begin{thebibliography}{57}
\providecommand{\natexlab}[1]{#1}
\providecommand{\url}[1]{\texttt{#1}}
\expandafter\ifx\csname urlstyle\endcsname\relax
  \providecommand{\doi}[1]{doi: #1}\else
  \providecommand{\doi}{doi: \begingroup \urlstyle{rm}\Url}\fi

\bibitem[Agrawal et~al.(2016)Agrawal, Batra, and Parikh]{agrawal2016analyzing}
Agrawal, A., Batra, D., and Parikh, D.
\newblock Analyzing the behavior of visual question answering models.
\newblock \emph{arXiv preprint arXiv:1606.07356}, 2016.

\bibitem[Arnab et~al.(2021)Arnab, Dehghani, Heigold, Sun, Lu{\v{c}}i{\'c}, and
  Schmid]{arnab2021vivit}
Arnab, A., Dehghani, M., Heigold, G., Sun, C., Lu{\v{c}}i{\'c}, M., and Schmid,
  C.
\newblock Vivit: A video vision transformer.
\newblock In \emph{Proceedings of the IEEE/CVF International Conference on
  Computer Vision}, pp.\  6836--6846, 2021.

\bibitem[Benton et~al.(2020)Benton, Finzi, Izmailov, and
  Wilson]{benton2020learning}
Benton, G., Finzi, M., Izmailov, P., and Wilson, A.~G.
\newblock Learning invariances in neural networks from training data.
\newblock \emph{Advances in neural information processing systems},
  33:\penalty0 17605--17616, 2020.

\bibitem[Bertasius et~al.(2021)Bertasius, Wang, and
  Torresani]{gberta_2021_ICML}
Bertasius, G., Wang, H., and Torresani, L.
\newblock Is space-time attention all you need for video understanding?
\newblock In \emph{Proceedings of the International Conference on Machine
  Learning (ICML)}, July 2021.

\bibitem[Buch et~al.(2022{\natexlab{a}})Buch, Eyzaguirre, Gaidon, Wu, Fei-Fei,
  and Niebles]{ATP}
Buch, S., Eyzaguirre, C., Gaidon, A., Wu, J., Fei-Fei, L., and Niebles, J.~C.
\newblock Revisiting the "video" in video-language understanding,
  2022{\natexlab{a}}.
\newblock URL \url{https://arxiv.org/abs/2206.01720}.

\bibitem[Buch et~al.(2022{\natexlab{b}})Buch, Eyzaguirre, Gaidon, Wu, Fei-Fei,
  and Niebles]{buch2022revisiting}
Buch, S., Eyzaguirre, C., Gaidon, A., Wu, J., Fei-Fei, L., and Niebles, J.~C.
\newblock Revisiting the" video" in video-language understanding.
\newblock In \emph{Proceedings of the IEEE/CVF Conference on Computer Vision
  and Pattern Recognition}, pp.\  2917--2927, 2022{\natexlab{b}}.

\bibitem[Chen et~al.(2020)Chen, Li, Yu, El~Kholy, Ahmed, Gan, Cheng, and
  Liu]{chen2020uniter}
Chen, Y.-C., Li, L., Yu, L., El~Kholy, A., Ahmed, F., Gan, Z., Cheng, Y., and
  Liu, J.
\newblock Uniter: Universal image-text representation learning.
\newblock In \emph{European conference on computer vision}, pp.\  104--120.
  Springer, 2020.

\bibitem[Contributors(2020)]{2020mmaction2}
Contributors, M.
\newblock Openmmlab's next generation video understanding toolbox and
  benchmark.
\newblock \url{https://github.com/open-mmlab/mmaction2}, 2020.

\bibitem[Dave et~al.(2022)Dave, Gupta, Rizve, and Shah]{dave2022tclr}
Dave, I., Gupta, R., Rizve, M.~N., and Shah, M.
\newblock Tclr: Temporal contrastive learning for video representation.
\newblock \emph{Computer Vision and Image Understanding}, 219:\penalty0 103406,
  2022.

\bibitem[Devlin et~al.(2018)Devlin, Chang, Lee, and Toutanova]{devlin2018bert}
Devlin, J., Chang, M.-W., Lee, K., and Toutanova, K.
\newblock Bert: Pre-training of deep bidirectional transformers for language
  understanding.
\newblock \emph{arXiv preprint arXiv:1810.04805}, 2018.

\bibitem[Diba et~al.(2021)Diba, Sharma, Safdari, Lotfi, Sarfraz, Stiefelhagen,
  and Van~Gool]{diba2021vi2clr}
Diba, A., Sharma, V., Safdari, R., Lotfi, D., Sarfraz, S., Stiefelhagen, R.,
  and Van~Gool, L.
\newblock Vi2clr: Video and image for visual contrastive learning of
  representation.
\newblock In \emph{Proceedings of the IEEE/CVF International Conference on
  Computer Vision}, pp.\  1502--1512, 2021.

\bibitem[Fang et~al.(2020)Fang, Gokhale, Banerjee, Baral, and
  Yang]{fang2020video2commonsense}
Fang, Z., Gokhale, T., Banerjee, P., Baral, C., and Yang, Y.
\newblock Video2commonsense: Generating commonsense descriptions to enrich
  video captioning.
\newblock \emph{arXiv preprint arXiv:2003.05162}, 2020.

\bibitem[Fire \& Zhu(2017)Fire and Zhu]{fire2017inferring}
Fire, A. and Zhu, S.-C.
\newblock Inferring hidden statuses and actions in video by causal reasoning.
\newblock In \emph{Proceedings of the IEEE Conference on Computer Vision and
  Pattern Recognition Workshops}, pp.\  48--56, 2017.

\bibitem[Han et~al.(2020)Han, Xie, and Zisserman]{han2020self}
Han, T., Xie, W., and Zisserman, A.
\newblock Self-supervised co-training for video representation learning.
\newblock \emph{Advances in Neural Information Processing Systems},
  33:\penalty0 5679--5690, 2020.

\bibitem[Hara et~al.(2017)Hara, Kataoka, and Satoh]{hara3dcnns}
Hara, K., Kataoka, H., and Satoh, Y.
\newblock Can spatiotemporal 3d cnns retrace the history of 2d cnns and
  imagenet?
\newblock \emph{arXiv preprint}, arXiv:1711.09577, 2017.

\bibitem[Jiang \& Han(2020)Jiang and Han]{HGA}
Jiang, P. and Han, Y.
\newblock Reasoning with heterogeneous graph alignment for video question
  answering.
\newblock \emph{Proceedings of the AAAI Conference on Artificial Intelligence},
  34\penalty0 (07):\penalty0 11109--11116, Apr. 2020.
\newblock \doi{10.1609/aaai.v34i07.6767}.
\newblock URL \url{https://ojs.aaai.org/index.php/AAAI/article/view/6767}.

\bibitem[Johnson et~al.(2017)Johnson, Hariharan, Van Der~Maaten, Fei-Fei,
  Lawrence~Zitnick, and Girshick]{johnson2017clevr}
Johnson, J., Hariharan, B., Van Der~Maaten, L., Fei-Fei, L., Lawrence~Zitnick,
  C., and Girshick, R.
\newblock Clevr: A diagnostic dataset for compositional language and elementary
  visual reasoning.
\newblock In \emph{Proceedings of the IEEE conference on computer vision and
  pattern recognition}, pp.\  2901--2910, 2017.

\bibitem[Johnson et~al.(2019)Johnson, Douze, and J{\'e}gou]{johnson2019billion}
Johnson, J., Douze, M., and J{\'e}gou, H.
\newblock Billion-scale similarity search with {GPUs}.
\newblock \emph{IEEE Transactions on Big Data}, 7\penalty0 (3):\penalty0
  535--547, 2019.

\bibitem[Kay et~al.(2017)Kay, Carreira, Simonyan, Zhang, Hillier,
  Vijayanarasimhan, Viola, Green, Back, Natsev, et~al.]{kay2017kinetics}
Kay, W., Carreira, J., Simonyan, K., Zhang, B., Hillier, C., Vijayanarasimhan,
  S., Viola, F., Green, T., Back, T., Natsev, P., et~al.
\newblock The kinetics human action video dataset.
\newblock \emph{arXiv preprint arXiv:1705.06950}, 2017.

\bibitem[Kipf \& Welling(2016)Kipf and Welling]{kipf2016semi}
Kipf, T.~N. and Welling, M.
\newblock Semi-supervised classification with graph convolutional networks.
\newblock \emph{arXiv preprint arXiv:1609.02907}, 2016.

\bibitem[Leen(1994)]{NIPS1994_7fa732b5}
Leen, T.
\newblock From data distributions to regularization in invariant learning.
\newblock In Tesauro, G., Touretzky, D., and Leen, T. (eds.), \emph{Advances in
  Neural Information Processing Systems}, volume~7. MIT Press, 1994.
\newblock URL
  \url{https://proceedings.neurips.cc/paper/1994/file/7fa732b517cbed14a48843d74526c11a-Paper.pdf}.

\bibitem[Lei et~al.(2018)Lei, Yu, Bansal, and Berg]{lei2018tvqa}
Lei, J., Yu, L., Bansal, M., and Berg, T.~L.
\newblock Tvqa: Localized, compositional video question answering.
\newblock \emph{arXiv preprint arXiv:1809.01696}, 2018.

\bibitem[Lei et~al.(2021{\natexlab{a}})Lei, Berg, and Bansal]{lei2021detecting}
Lei, J., Berg, T.~L., and Bansal, M.
\newblock Detecting moments and highlights in videos via natural language
  queries.
\newblock \emph{Advances in Neural Information Processing Systems},
  34:\penalty0 11846--11858, 2021{\natexlab{a}}.

\bibitem[Lei et~al.(2021{\natexlab{b}})Lei, Berg, and Bansal]{qvhighlights}
Lei, J., Berg, T.~L., and Bansal, M.
\newblock Qvhighlights: Detecting moments and highlights in videos via natural
  language queries, 2021{\natexlab{b}}.
\newblock URL \url{https://arxiv.org/abs/2107.09609}.

\bibitem[Lei et~al.(2021{\natexlab{c}})Lei, Li, Zhou, Gan, Berg, Bansal, and
  Liu]{lei2021less}
Lei, J., Li, L., Zhou, L., Gan, Z., Berg, T.~L., Bansal, M., and Liu, J.
\newblock Less is more: Clipbert for video-and-language learning via sparse
  sampling.
\newblock In \emph{Proceedings of the IEEE/CVF Conference on Computer Vision
  and Pattern Recognition}, pp.\  7331--7341, 2021{\natexlab{c}}.

\bibitem[Li et~al.(2021{\natexlab{a}})Li, Wang, Zhang, Li, Keutzer, Darrell,
  and Zhao]{li2021learning}
Li, B., Wang, Y., Zhang, S., Li, D., Keutzer, K., Darrell, T., and Zhao, H.
\newblock Learning invariant representations and risks for semi-supervised
  domain adaptation.
\newblock In \emph{Proceedings of the IEEE/CVF Conference on Computer Vision
  and Pattern Recognition}, pp.\  1104--1113, 2021{\natexlab{a}}.

\bibitem[Li et~al.(2020)Li, Chen, Cheng, Gan, Yu, and Liu]{li2020hero}
Li, L., Chen, Y.-C., Cheng, Y., Gan, Z., Yu, L., and Liu, J.
\newblock Hero: Hierarchical encoder for video+ language omni-representation
  pre-training.
\newblock \emph{arXiv preprint arXiv:2005.00200}, 2020.

\bibitem[Li et~al.(2021{\natexlab{b}})Li, Lei, Gan, Yu, Chen, Pillai, Cheng,
  Zhou, Wang, Wang, et~al.]{li2021value}
Li, L., Lei, J., Gan, Z., Yu, L., Chen, Y.-C., Pillai, R., Cheng, Y., Zhou, L.,
  Wang, X.~E., Wang, W.~Y., et~al.
\newblock Value: A multi-task benchmark for video-and-language understanding
  evaluation.
\newblock \emph{arXiv preprint arXiv:2106.04632}, 2021{\natexlab{b}}.

\bibitem[Li et~al.(2022{\natexlab{a}})Li, Wang, Xiao, and Chua]{HGA+EIGV}
Li, Y., Wang, X., Xiao, J., and Chua, T.-S.
\newblock Equivariant and invariant grounding for video question answering,
  2022{\natexlab{a}}.
\newblock URL \url{https://arxiv.org/abs/2207.12783}.

\bibitem[Li et~al.(2022{\natexlab{b}})Li, Wang, Xiao, and
  Chua]{li2022equivariant}
Li, Y., Wang, X., Xiao, J., and Chua, T.-S.
\newblock Equivariant and invariant grounding for video question answering.
\newblock In \emph{Proceedings of the 30th ACM International Conference on
  Multimedia}, pp.\  4714--4722, 2022{\natexlab{b}}.

\bibitem[Li et~al.(2022{\natexlab{c}})Li, Wang, Xiao, Ji, and Chua]{Li_IGV}
Li, Y., Wang, X., Xiao, J., Ji, W., and Chua, T.-S.
\newblock Invariant grounding for video question answering.
\newblock In \emph{Proceedings of the IEEE/CVF Conference on Computer Vision
  and Pattern Recognition (CVPR)}, pp.\  2928--2937, June 2022{\natexlab{c}}.

\bibitem[Lin et~al.(2019)Lin, Gan, and Han]{lin2019tsm}
Lin, J., Gan, C., and Han, S.
\newblock Tsm: Temporal shift module for efficient video understanding.
\newblock In \emph{Proceedings of the IEEE/CVF International Conference on
  Computer Vision}, pp.\  7083--7093, 2019.

\bibitem[Lin et~al.(2022)Lin, Li, Lin, Ahmed, Gan, Liu, Lu, and
  Wang]{lin2021end-to-end}
Lin, K., Li, L., Lin, C.-C., Ahmed, F., Gan, Z., Liu, Z., Lu, Y., and Wang, L.
\newblock Swinbert: End-to-end transformers with sparse attention for video
  captioning.
\newblock In \emph{CVPR}, 2022.

\bibitem[Lopez-Paz et~al.(2017)Lopez-Paz, Nishihara, Chintala, Scholkopf, and
  Bottou]{lopez2017discovering}
Lopez-Paz, D., Nishihara, R., Chintala, S., Scholkopf, B., and Bottou, L.
\newblock Discovering causal signals in images.
\newblock In \emph{Proceedings of the IEEE conference on computer vision and
  pattern recognition}, pp.\  6979--6987, 2017.

\bibitem[Lu et~al.(2019)Lu, Batra, Parikh, and Lee]{lu2019vilbert}
Lu, J., Batra, D., Parikh, D., and Lee, S.
\newblock Vilbert: Pretraining task-agnostic visiolinguistic representations
  for vision-and-language tasks.
\newblock \emph{Advances in neural information processing systems}, 32, 2019.

\bibitem[Miech et~al.(2020)Miech, Alayrac, Smaira, Laptev, Sivic, and
  Zisserman]{miech2020end}
Miech, A., Alayrac, J.-B., Smaira, L., Laptev, I., Sivic, J., and Zisserman, A.
\newblock End-to-end learning of visual representations from uncurated
  instructional videos.
\newblock In \emph{Proceedings of the IEEE/CVF Conference on Computer Vision
  and Pattern Recognition}, pp.\  9879--9889, 2020.

\bibitem[Misra \& Maaten(2020)Misra and Maaten]{Misra_2020_CVPR}
Misra, I. and Maaten, L. v.~d.
\newblock Self-supervised learning of pretext-invariant representations.
\newblock In \emph{Proceedings of the IEEE/CVF Conference on Computer Vision
  and Pattern Recognition (CVPR)}, June 2020.

\bibitem[Oord et~al.(2018)Oord, Li, and Vinyals]{oord2018representation}
Oord, A. v.~d., Li, Y., and Vinyals, O.
\newblock Representation learning with contrastive predictive coding.
\newblock \emph{arXiv preprint arXiv:1807.03748}, 2018.

\bibitem[Qi et~al.(2020)Qi, Niu, Huang, and Zhang]{qi2020two}
Qi, J., Niu, Y., Huang, J., and Zhang, H.
\newblock Two causal principles for improving visual dialog.
\newblock In \emph{Proceedings of the IEEE/CVF conference on computer vision
  and pattern recognition}, pp.\  10860--10869, 2020.

\bibitem[Radford et~al.(2021{\natexlab{a}})Radford, Kim, Hallacy, Ramesh, Goh,
  Agarwal, Sastry, Askell, Mishkin, Clark, Krueger, and Sutskever]{clip}
Radford, A., Kim, J.~W., Hallacy, C., Ramesh, A., Goh, G., Agarwal, S., Sastry,
  G., Askell, A., Mishkin, P., Clark, J., Krueger, G., and Sutskever, I.
\newblock Learning transferable visual models from natural language
  supervision, 2021{\natexlab{a}}.
\newblock URL \url{https://arxiv.org/abs/2103.00020}.

\bibitem[Radford et~al.(2021{\natexlab{b}})Radford, Kim, Hallacy, Ramesh, Goh,
  Agarwal, Sastry, Askell, Mishkin, Clark, et~al.]{radford2021learning}
Radford, A., Kim, J.~W., Hallacy, C., Ramesh, A., Goh, G., Agarwal, S., Sastry,
  G., Askell, A., Mishkin, P., Clark, J., et~al.
\newblock Learning transferable visual models from natural language
  supervision.
\newblock In \emph{International Conference on Machine Learning}, pp.\
  8748--8763. PMLR, 2021{\natexlab{b}}.

\bibitem[Sohn \& Lee(2012)Sohn and Lee]{sohn2012learning}
Sohn, K. and Lee, H.
\newblock Learning invariant representations with local transformations.
\newblock \emph{arXiv preprint arXiv:1206.6418}, 2012.

\bibitem[Storks et~al.(2019)Storks, Gao, and Chai]{storks2019commonsense}
Storks, S., Gao, Q., and Chai, J.~Y.
\newblock Commonsense reasoning for natural language understanding: A survey of
  benchmarks, resources, and approaches.
\newblock \emph{arXiv preprint arXiv:1904.01172}, pp.\  1--60, 2019.

\bibitem[Sun et~al.(2019)Sun, Myers, Vondrick, Murphy, and
  Schmid]{sun2019videobert}
Sun, C., Myers, A., Vondrick, C., Murphy, K., and Schmid, C.
\newblock Videobert: A joint model for video and language representation
  learning.
\newblock In \emph{Proceedings of the IEEE/CVF International Conference on
  Computer Vision}, pp.\  7464--7473, 2019.

\bibitem[Vaswani et~al.(2017)Vaswani, Shazeer, Parmar, Uszkoreit, Jones, Gomez,
  Kaiser, and Polosukhin]{vaswani2017attention}
Vaswani, A., Shazeer, N., Parmar, N., Uszkoreit, J., Jones, L., Gomez, A.~N.,
  Kaiser, {\L}., and Polosukhin, I.
\newblock Attention is all you need.
\newblock \emph{Advances in neural information processing systems}, 30, 2017.

\bibitem[Wang et~al.(2019)Wang, Wu, Chen, Li, Wang, and Wang]{wang2019vatex}
Wang, X., Wu, J., Chen, J., Li, L., Wang, Y.-F., and Wang, W.~Y.
\newblock Vatex: A large-scale, high-quality multilingual dataset for
  video-and-language research.
\newblock In \emph{Proceedings of the IEEE/CVF International Conference on
  Computer Vision}, pp.\  4581--4591, 2019.

\bibitem[Wang et~al.(2020)Wang, Menkovski, Wang, Du, and
  Pechenizkiy]{wang2020causal}
Wang, Y., Menkovski, V., Wang, H., Du, X., and Pechenizkiy, M.
\newblock Causal discovery from incomplete data: a deep learning approach.
\newblock \emph{arXiv preprint arXiv:2001.05343}, 2020.

\bibitem[Wörtwein et~al.(2022)Wörtwein, Sheeber, Allen, Cohn, and
  Morency]{residualmma}
Wörtwein, T., Sheeber, L.~B., Allen, N., Cohn, J.~F., and Morency, L.-P.
\newblock Beyond additive fusion: Learning non-additive multimodal
  interactions.
\newblock In \emph{Findings of the Association for Computational Linguistics:
  EMNLP 2022}, 2022.
\newblock URL
  \url{https://github.com/twoertwein/MultimodalResidualOptimization/blob/main/paper.pdf}.

\bibitem[Xiao et~al.(2021{\natexlab{a}})Xiao, Shang, Yao, and Chua]{nextQA}
Xiao, J., Shang, X., Yao, A., and Chua, T.-S.
\newblock Next-qa:next phase of question-answering to explaining temporal
  actions, 2021{\natexlab{a}}.
\newblock URL \url{https://arxiv.org/abs/2105.08276}.

\bibitem[Xiao et~al.(2022)Xiao, Zhou, Chua, and Yan]{xiao2022video}
Xiao, J., Zhou, P., Chua, T.-S., and Yan, S.
\newblock Video graph transformer for video question answering.
\newblock In \emph{European Conference on Computer Vision}, pp.\  39--58.
  Springer, 2022.

\bibitem[Xiao et~al.(2021{\natexlab{b}})Xiao, Shen, Zhen, Shao, and
  Snoek]{pmlr-v139-xiao21a}
Xiao, Z., Shen, J., Zhen, X., Shao, L., and Snoek, C.
\newblock A bit more bayesian: Domain-invariant learning with uncertainty.
\newblock In Meila, M. and Zhang, T. (eds.), \emph{Proceedings of the 38th
  International Conference on Machine Learning}, volume 139 of
  \emph{Proceedings of Machine Learning Research}, pp.\  11351--11361. PMLR,
  18--24 Jul 2021{\natexlab{b}}.
\newblock URL \url{https://proceedings.mlr.press/v139/xiao21a.html}.

\bibitem[Xu et~al.(2016)Xu, Mei, Yao, and Rui]{xu2016msr}
Xu, J., Mei, T., Yao, T., and Rui, Y.
\newblock Msr-vtt: A large video description dataset for bridging video and
  language.
\newblock In \emph{Proceedings of the IEEE conference on computer vision and
  pattern recognition}, pp.\  5288--5296, 2016.

\bibitem[Yu et~al.(2018)Yu, Kim, and Kim]{yu2018joint}
Yu, Y., Kim, J., and Kim, G.
\newblock A joint sequence fusion model for video question answering and
  retrieval.
\newblock In \emph{Proceedings of the European Conference on Computer Vision
  (ECCV)}, pp.\  471--487, 2018.

\bibitem[Zhang et~al.(2016)Zhang, Goyal, Summers-Stay, Batra, and
  Parikh]{zhang2016yin}
Zhang, P., Goyal, Y., Summers-Stay, D., Batra, D., and Parikh, D.
\newblock Yin and yang: Balancing and answering binary visual questions.
\newblock In \emph{Proceedings of the IEEE conference on computer vision and
  pattern recognition}, pp.\  5014--5022, 2016.

\bibitem[Zhou et~al.(2018)Zhou, Andonian, Oliva, and
  Torralba]{zhou2018temporal}
Zhou, B., Andonian, A., Oliva, A., and Torralba, A.
\newblock Temporal relational reasoning in videos.
\newblock In \emph{Proceedings of the European conference on computer vision
  (ECCV)}, pp.\  803--818, 2018.

\bibitem[Zhou et~al.(2020)Zhou, Wang, Liu, Hu, and Zhang]{zhou2020more}
Zhou, Y., Wang, M., Liu, D., Hu, Z., and Zhang, H.
\newblock More grounded image captioning by distilling image-text matching
  model.
\newblock In \emph{Proceedings of the IEEE/CVF conference on computer vision
  and pattern recognition}, pp.\  4777--4786, 2020.

\bibitem[Zhu \& Yang(2020)Zhu and Yang]{zhu2020actbert}
Zhu, L. and Yang, Y.
\newblock Actbert: Learning global-local video-text representations.
\newblock In \emph{Proceedings of the IEEE/CVF conference on computer vision
  and pattern recognition}, pp.\  8746--8755, 2020.

\end{thebibliography}
\bibliographystyle{icml2022}
\newpage
\appendix
\section{Appendix}
\label{appendix}

\subsection{EIGV scene intervener and causal disruptor modules}
\label{app:eigv-eq}
The scene intervener creates the $q^*$, $a^*$, $c^*$ and $t^*$ which are the casual factors - Question, Answer, Causal scene and Complement scene respectively and $c^*$ and $t^*$ are from video $v$. They are obtained as linear interpolation between two data points on their causal factor - $C,\mathcal{Q}$ and $\mathcal{A}$ with different mixing ratios $\lambda_0\sim$Beta($\alpha, \alpha$) for causal scene mixup and $\lambda_1\sim$U(0,1) for complement scene mixup as shown below.
\begin{equation}
c^* = \lambda_0 \hat{c} + (1-\lambda_0)\hat{c}'
\end{equation}
\begin{equation}
q^* = \lambda_0 \hat{q} + (1-\lambda_0)\hat{q}'
\end{equation}
\begin{equation}
a^* = \lambda_0\hat{a} + (1-\lambda_0)\hat{a}'
\end{equation}
\begin{equation}
t^* = \lambda_1\hat{t} + (1-\lambda_1)\hat{t}'
\end{equation}
where $\hat{c}'$, $\hat{q}'$, $\hat{a}'$ and $\hat{t}'$ come from a second data sample ($\mathcal{V}'$) used for the mixup.
The causal disruptor combines both these elements into a contrastive loss where a positive video $\mathcal{V}^+$ is obtained by disrupting $\mathcal{V}^*$ on the complement by substituting with an item randomly sampled from a memory bank $\pi$. Negative counterpart $\mathcal{V}^-$ obtained using E-intervention on $\mathcal{V}^*$ by substituting question-critical causal part. It also creates linguistic alternatives as negative sample by disrupting $(\mathcal{V}^*, q^*)$ to $(\mathcal{V}^*, q_r)$ where $q_r$ is a random sample question. 
\begin{equation}
\label{eq:anchor_eigv}
a = \mathcal{F}_{\hat{A}}(\mathcal{V}^*,q^*),
\end{equation}
\begin{equation}
\label{eq:positive_eigv}
a^+ = \mathcal{F}_{\hat{A}}(\mathcal{V}^+,q^*),
\end{equation}
\begin{equation}
\label{eq:negative_eigv}
a^- = \mathcal{F}_{\hat{A}}[concat((\mathcal{V}^-,q^*),(\mathcal{V}^*,q_r))]
\end{equation}
The contrastive loss is then given by InfoNCE as:
\begin{equation}
\mathcal{L}_{CL} = -log(\frac{exp(a^Ta^+)}{exp(a^Ta^+)+\sum_n^Nexp(a^Ta^-_n)})
\end{equation}
where N is the number of negative samples, and $a^-_n$ denotes answer generated by one negative sample. This combined with ERM objective, we have:
\begin{equation}
\mathcal{L}_{EIGV} = \mathbb{E}_{(v, q, a) \in \mathcal{O}}\mathcal{L}_{ERM} + \beta\mathcal{L}_{CL}
\end{equation}
where $\mathcal{O}$ is set of training instances, $\beta$ is a hyperparameter to balance CL strength. 

\subsection{Video Graph Transformer} \label{vgt}
Contrary to the single frame/single clip approaches such as \cite{li2021learning,buch2022revisiting}, or identifying holistic regions of the video for answering questions~\cite{yu2018joint,zhu2020actbert,lei2018tvqa}, \cite{xiao2022video} proposes to use the objects and their \textit{spatio-temporal} relationships, which vary across time, in addition to using the video, questions and answer choices as the modalities of information. Towards this end they proposed a Dynamic Graph Transformer based VideoQA model (VGT) with emphasis on capturing the dynamics of objects and relations for better reasoning, and separate transformers for text and video modalities to better encode the information from the corresponding modalities. The different components of the model are as follows - \textbf{video graph representation}: they split a video into \textit{n} clips and sample evenly distributed frames for each clip to identify the objects and the spatial locations of the objects in each frame of a clip. The consistency of objects across frames is obtained through IoU overlap and similarity of the objects. Once the objects are fixed for a clip (across frames), the graph is constructed at each frame by computing edge scores as 
\begin{align}    
E_{i,j} = \sigma(\phi_{W_{ak}}(\mathcal{F}_{o_i})\odot\phi_{W_{av}}(\mathcal{F}_{o_j}))
\end{align}
where \textit{i, j} are the objects, $\mathcal{F}_{o_i}$ is the encoding of object \textit{i}, which is made up of ROI pooling of visual and spatial locations, and the $\phi$ matrices are the transformations learnt in the model. A \textbf{temporal graph transformer} modifies the edge weights as the node representations propagate through transformer layers, i.e., the object representations enhance by \textit{attending} to the other objects and the edge weights are updated accordingly. A \textbf{spatial graph convolution} operation uses the graph attention convolution \cite{kipf2016semi} to encode spatial relations, and a final \textbf{hierarchical aggregation} layer, which is a weighted representation of the nodes in the graphs of the frames. Finally a mean pooling layer aggregates the clips to obtain the final \textbf{video relational representation}.

After encoding text modality using the $\mathcal{E}_{q}, \mathcal{E}_{c}$ modules, the cross modal interactions are captured through additional attention modules. The final answer is generated based on the multimodal \textit{qv} and answer representations.
\begin{align}
E_{qv} = E_v + \sum_{m=1}^{M} \beta_m E_{qm} \text{ where } \beta = \sigma(E_v\odot E_q) \\
s_i = \mathcal{F}_{qv}(E_{qv})\odot \mathcal{F}_{C}(\mathcal{E}_{c}(\mathcal{C}_i)) \text{ for } i\in [1, 5]
\end{align}
where $E_v$ is the video representation obtained from graph transformer and $E_{qm}$ is the representation for $m^{th}$ token in the question, $\mathcal{F}_{qv}, \mathcal{F}_{C}$ are transformations of \textit{question-video} and \textit{candidate answer} representations. The final objective is a simple ERM softmax loss as defined in Equation \ref{eq:erm}.

\begin{figure}[ht]
    \centering
    \includegraphics[width=1.0\columnwidth]{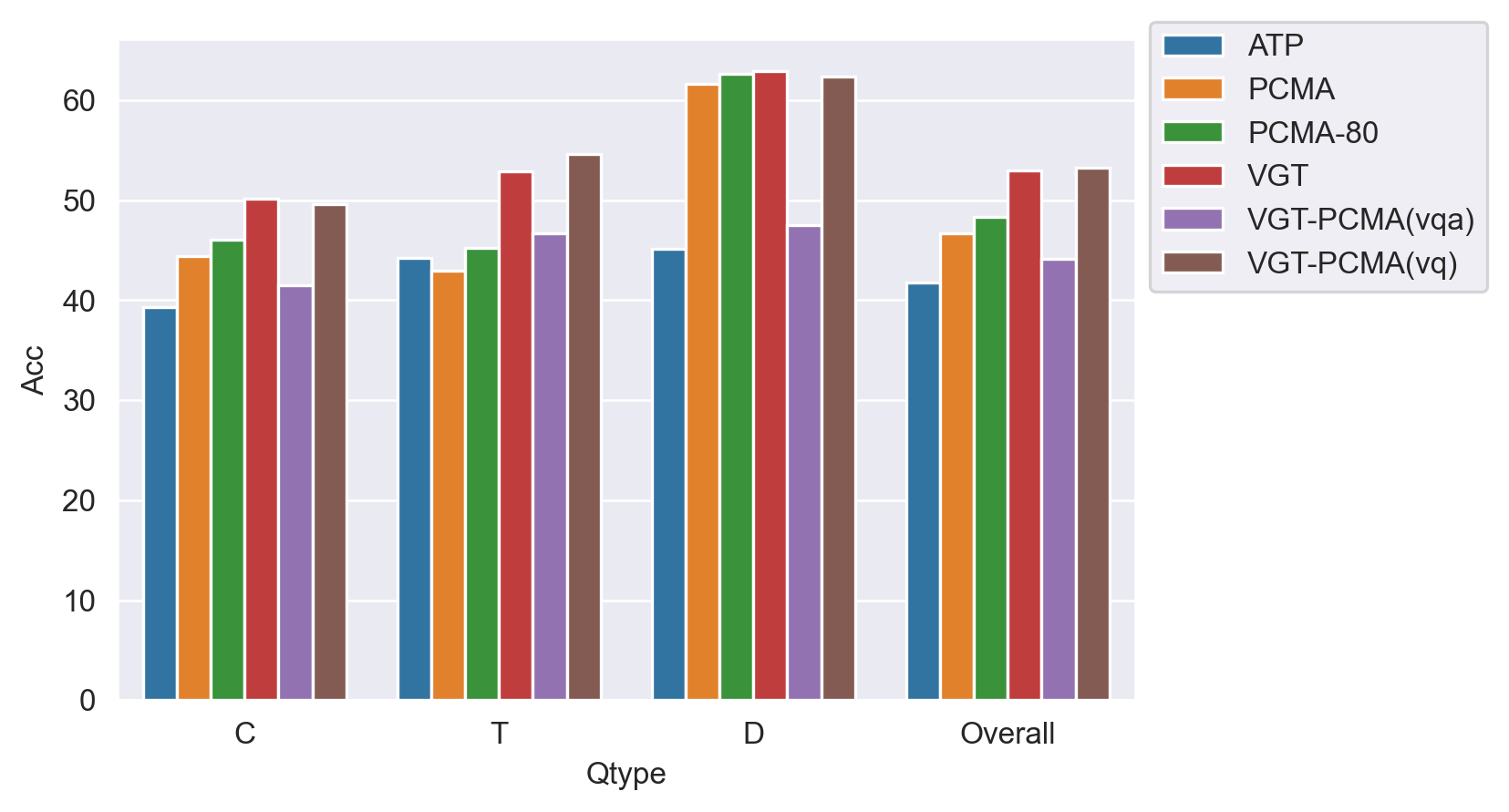}
    \caption{Comparing performance with ATP, VGT and Sampled Inputs}
    \label{fig:cma_acc}
\end{figure}
\subsection{Action Grounding Ablations}

In Table \ref{action_ground_ablations}, list of acronyms:\\
N: Number of Nearest Neighbors \\
NumFrames: Number of Frames \\
TF: Text Features (Either from Next or using Captions and/or actions) \\
VF: Video Features (Taken from Grounded Video or Next Dataset) \\
Bef\&Aft: Before and After \\
Acc\_C: Accuracy of Causal \\
Acc\_T: Accuracy of Temporal \\
Acc\_D: Accuracy of Descriptive \\

From the table, we can see that the combination of grounded video features, next text features as well as using 3 nearest neighbors from FAISS perform the best. This is closely followed by using 5 neighbors and using no neighbors at all. This shows that using FAISS increases the generalization performance. In addition, across all categories, we see that the approaches that utilize FAISS outperform vanilla models across all categories.

\subsection{Qualitative Analysis of Action Grounding}
\begin{figure*}[ht]
    \centering
    \includegraphics[scale=0.35]{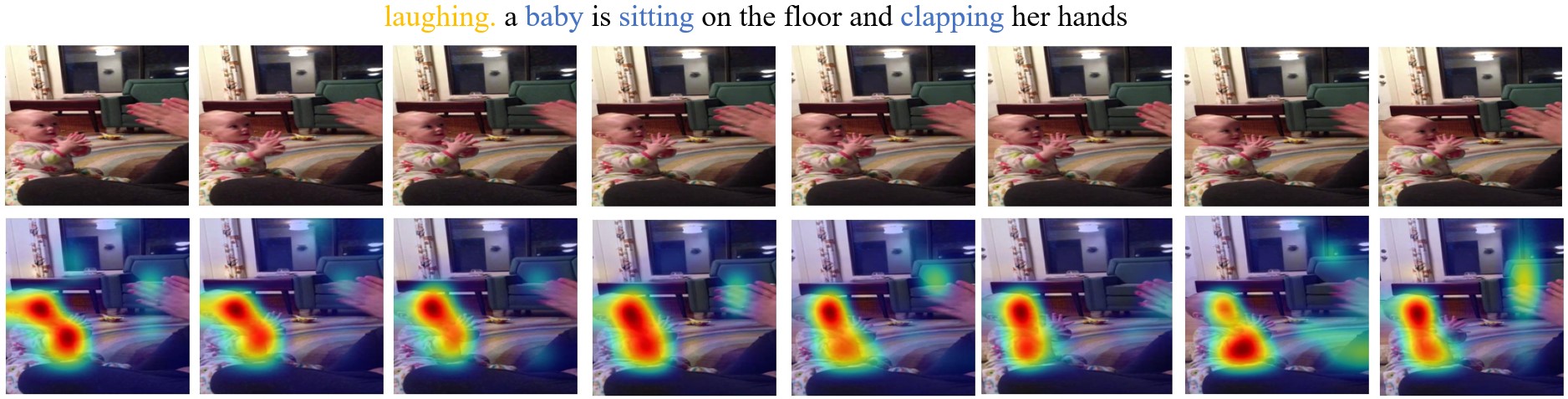}
    \caption{The model is able to match the most relevant frame with the relevant parts of the text}
    \label{fig:action_qual_eg2}
\end{figure*}

\begin{figure*}[ht]
    \centering
    \includegraphics[scale=0.35]{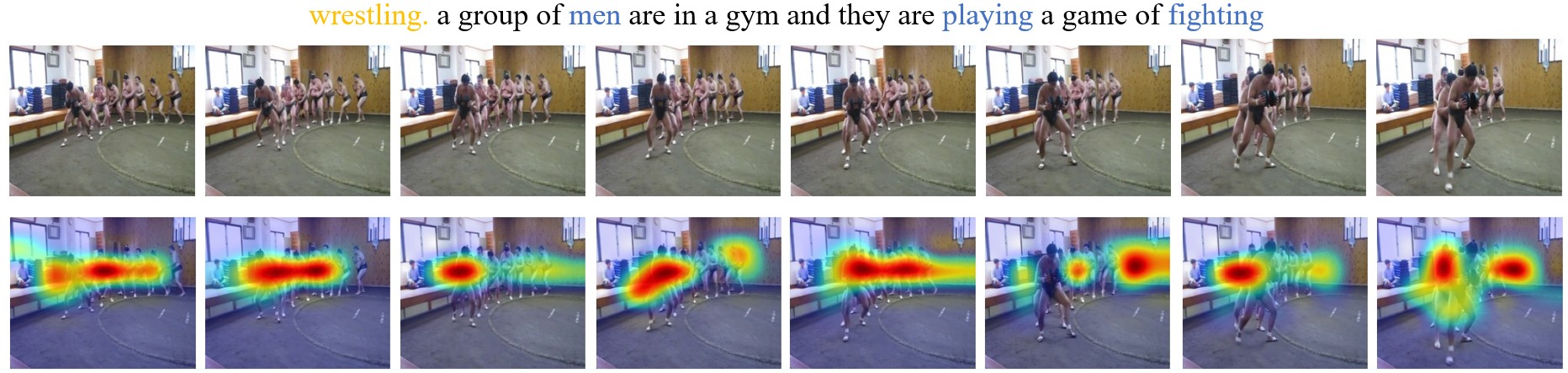}
    \caption{The model is able to match the most relevant frame with the relevant parts of the text}
    \label{fig:action_qual_eg3}
\end{figure*}

In fig~\ref{fig:action_qual_eg1}, \ref{fig:action_qual_eg2} and \ref{fig:action_qual_eg3} we see that the combination of actions in conjunction with information of the objects helps ground the interactions happening in the video.

\begin{table*}[bt]
\scalebox{.8}{
\begin{tabular}{|c|c|c|c|c|c|c|c|c|c|c|c|c|c|c|c|}
\hline
N & NumFrames & TF & VF & Val & Test & Why & How & Bef\&Aft & When & Cnt & Loc & Other & Acc\_C & Acc\_T & Acc\_D \\
\hline
3 & 32 & next & ground & 55.94 & \textbf{54.86} & \textbf{54.46} & 49.19 & 51.01 & 57.34 & 46.89 & 80.33 & 56.67 & \textbf{53.09} & 53.78 & 62.56 \\
5 & 32 & next & ground & 55.54 & 54.79 & 53.77 & 49.36 & \textbf{51.94} & \textbf{57.94} & 44.41 & 80.12 & 57.17 & 52.62 & \textbf{54.57} & 62.14 \\
- & 32 & next & ground & 56 & 54.59 & 53.59 & 49.96 & 49.73 & 56.22 & 48.45 & \textbf{83.64} & \textbf{58} & 52.64 & 52.58 & \textbf{64.63} \\
3 & 16 & next & ground & 55.7 & 54.59 & 53.56 & \textbf{50.38} & 51.01 & 56.74 & 47.83 & 80.54 & 56 & 52.73 & 53.52 & 62.56 \\
5 & 16 & next & ground & 55.6 & 54.46 & 53.53 & 47.74 & 50.87 & 57.17 & 49.38 & 82.19 & 56.83 & 52.02 & 53.63 & 63.84 \\
- & 16 & next & ground & \textbf{56.12} & 54.44 & 53.11 & 49.36 & 50.54 & 56.82 & 49.69 & 82.19 & 57 & 52.13 & 53.29 & 63.99 \\
\hline
- & 16 & next & next & 53.28 & 52.41 & 50.05 & 48.34 & 50.34 & 53.91 & 48.14 & 79.3 & 56.33 & 49.6 & 51.9 & 62.35 \\
- & 32 & caption & ground & 50.58 & 50.43 & 49.35 & 43.56 & 46.11 & 52.45 & 48.76 & 80.75 & 53.17 & 47.85 & 48.89 & 61.64 \\
- & 32 & action + caption & ground & 50 & 49.65 & 48.6 & 42.71 & 46.25 & 49.1 & \textbf{50} & 80.12 & 53.83 & 47.07 & 47.5 & 61.99 \\
- & 16 & caption & next & 48.96 & 48.88 & 47.1 & 42.54 & 46.58 & 49.87 & 48.45 & 77.43 & 52.17 & 45.91 & 48.02 & 60 \\
- & 16 & action + caption & next & 49.26 & 48.83 & 48.3 & 41.01 & 45.91 & 50.39 & 45.96 & 76.4 & 50.67 & 46.4 & 47.87 & 58.43 \\
- & 16 & action & next & 20.32 & 20.78 & 20.52 & 19.01 & 21.31 & 22.4 & 24.53 & 19.25 & 20.5 & 20.12 & 21.79 & 21 \\
- & 16 & action & ground & 20.64 & 19.79 & 19.8 & 19.18 & 21.38 & 19.48 & 20.19 & 15.53 & 20.83 & 19.64 & 20.55 & 18.86 \\
\hline
\end{tabular}
}
\caption{Ablation study for Action Grounding}
\label{action_ground_ablations}
\end{table*}

\begin{figure*}[]
    \centering
    \includegraphics[scale=0.25]{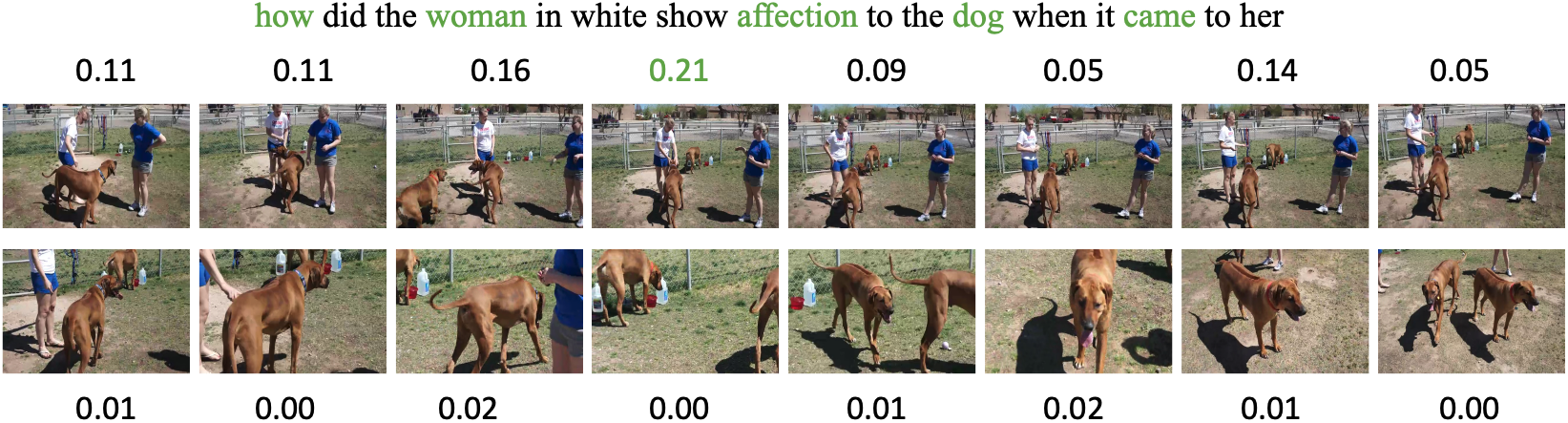}
    \caption{Qualitative example of PCMA module. The model is able to match the most relevant frame with the relevant parts of the text}
    \label{fig:cma_qual_eg}
\end{figure*}
\subsection{Smart Sampling}

\begin{figure*}[ht]
    \textcolor{black}{Q1: why is the girl hitting the box that she is sitting on?}
    \textcolor[HTML]{3C8031}{Ans.1: Playing the instrument}\\
    \textcolor{black}{Q2: why did the lady in beige put the pooh bear onto the floor?} 
    \textcolor[HTML]{2D2F92}{Ans.2: playing with the baby}\vspace{3mm}\\
    \includegraphics[width=1.0\textwidth]{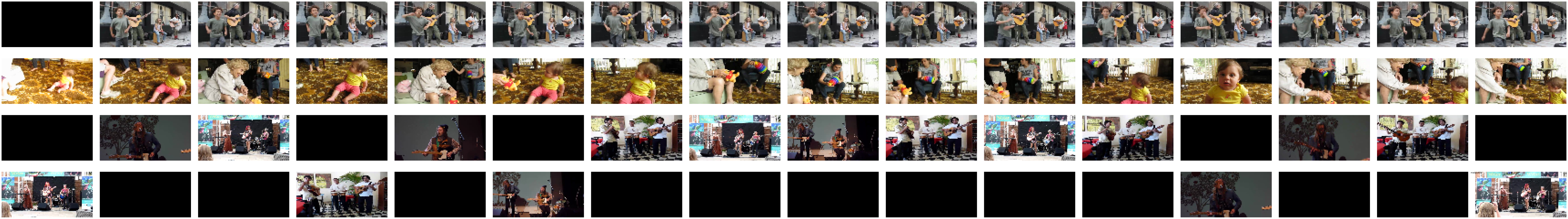}
    \caption{EIGV+MRI qualitative example 1. \textit{Top:} Two questions and the corresponding ground truth answers for two videos shown in rows 1 and 2. \textit{Bottom:} Row 1 and row 2 corresponds to 16 frames of videos that are mixed for with mixup factor $\lambda_0$ for intervention. Rows 3 and 4 show the frames that are replaced in the $\lambda_0$-mixed video; black indicates that original frames from the mixed video are retained. Row 3 corresponds to invariant intervention and row 4 corresponds to equivariant intervention. Note that rows 3 and 4 are complimentary. Also, the interventions are created for replacing the closest the frames in the video 1 for type 2 intervention in MRI(ref to Section~\ref{methodology}). Notice how close the intervening components are to the video 1. The original video and the intervention contain multiple people playing guitar instruments. Please zoom in for clear details.}
    \label{fig:faiss_qual_eg1}
\end{figure*}
\begin{figure*}[h]
    \textcolor{black}{Q1: what does the boy hold in his hands as he sits on the chair?}
    \textcolor[HTML]{3C8031}{Ans.1: Book}\\
    \textcolor{black}{Q2: how did the man in green bends the wood?} 
    \textcolor[HTML]{2D2F92}{Ans.2: with his hand}\vspace{3mm}\\
    \includegraphics[height=4cm,width=1.0\textwidth]{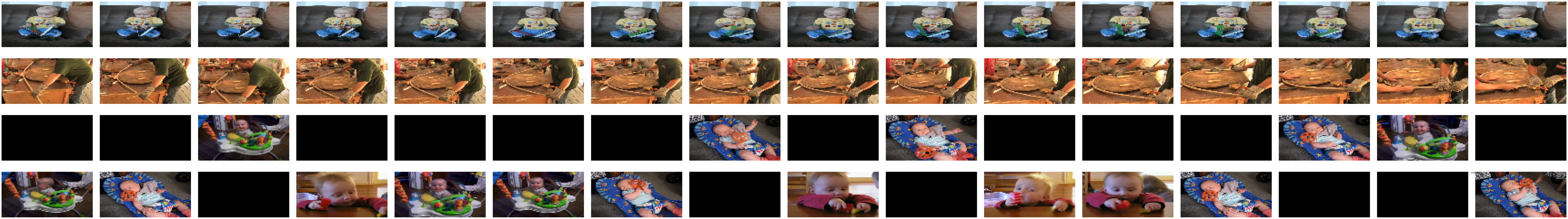}
    \caption{EIGV+MRI qualitative example 2. \textit{Top:} Two questions and the corresponding ground truth answers for two videos shown in rows 1 and 2. \textit{Bottom:} Row 1 and row 2 corresponds to 16 frames of videos that are mixed for with mixup factor $\lambda_0$ for intervention. Rows 3 and 4 show the frames that are replaced in the $\lambda_0$-mixed video; black indicates that original frames from the mixed video are retained. Row 3 corresponds to invariant intervention and row 4 corresponds to equivariant intervention. Note that rows 3 and 4 are complimentary. Also, the interventions are created for replacing the closest the frames in the video 1 for type 2 intervention in MRI(ref to Section~\ref{methodology}). Notice how close the intervening components are to the video 1. The original video and the intervention contain toddlers playing with items in their hands. Please zoom in for clear details.}
    \label{fig:faiss_qual_eg2}
\end{figure*}
\end{document}